\documentclass[letterpaper]{article} 
\usepackage[preprint]{aaai2027}
\usepackage[hyphens]{url}  
\usepackage{graphicx} 
\usepackage{natbib}  
\usepackage{caption} 
\usepackage{tabularx}
\usepackage{array}
\usepackage{amssymb}

\usepackage{algorithm}
\usepackage{booktabs}
\usepackage{multirow}
\usepackage{makecell}
\usepackage{threeparttable}
\usepackage{graphicx}
\usepackage{amssymb}
\usepackage{amsthm}

\usepackage[table]{xcolor}
\definecolor{raggreen}{RGB}{240,250,240}      
\definecolor{flatblue}{RGB}{235,247,252}      
\definecolor{structpink}{RGB}{252,238,244}    
\definecolor{ourspeach}{RGB}{255,238,221}  
\definecolor{trainpurple}{RGB}{244,240,252}   

\newcommand{\vDashSep}{%
  \hspace{2pt}%
  {\color{black!45}%
   \vrule width 0.4pt
   height .65\ht\strutbox
   depth .65\dp\strutbox}%
  \hspace{2pt}%
}

\usepackage{newfloat}
\usepackage{listings}
\usepackage{amsmath}

\DeclareCaptionStyle{ruled}{labelfont=normalfont,labelsep=colon,strut=off} 
\floatstyle{ruled}
\newfloat{listing}{tb}{lst}{}
\floatname{listing}{Listing}

\usepackage{booktabs}
\usepackage{algorithm}
\usepackage{algpseudocode}

\title{CoEvo-Mem: Co-Evolving Retrieval Policy and Memory Bank \\ for LLM Agents}
\author{
    Bowen Ye\textsuperscript{\rm 1,2},
    Yongchao Xu\textsuperscript{\rm 1,3},
    Zhijian Li\textsuperscript{\rm 1},
    Xiang Yin\textsuperscript{\rm 2},\\
    Junkai Ma\textsuperscript{\rm 1}\corresponding\equalcontrib,
    Wenzhao Li\textsuperscript{\rm 1}\equalcontrib,
}
\affiliations{
    \textsuperscript{\rm 1}Alibaba Group, Hangzhou, China \\
    \textsuperscript{\rm 2}School of Automation and Intelligent Sensing, Shanghai Jiao Tong University, Shanghai 200240, China\\
    \textsuperscript{\rm 3}University of Science and Technology of China, Hefei, China\\

}

\begin{document}

\maketitle

\begin{abstract}
As memories accumulate across tasks and sessions, the performance of long-term LLM agents depends jointly on query-specific retrieval and continual memory refinement. However, existing methods typically optimize either memory access, through iterative query refinement or adaptive retrieval policies, or memory evolution such as structural update. This separation overlooks a fundamental feedback loop: retrieval determines which memories receive usage signals, while updated memory bank reshape future retrieval. We propose \textbf{CoEvo-Mem}, a closed-loop framework for co-evolving the retrieval policy and memory bank. For each query, a frozen LLM generates route-specific query rewrites and a routing prior, which a lightweight residual router corrects online. The retrieved context serves as the coupling interface between the two learning processes: task outcomes assign credit to routing decisions, while trajectory-conditioned feedback updates memory values and graph relations. These updates alter how memories are ranked and selected for subsequent queries, thereby closing the feedback loop. To mitigate coupling induced non-stationarity, CoEvo-Mem alternates between updating the router with the memory bank fixed and evolving the memory bank with the retrieval policy fixed. Across seven diverse benchmarks, \textbf{CoEvo-Mem} achieves state-of-the-art performance, demonstrating the importance of retrieval-memory coevolution.
\end{abstract}

\section{Introduction}
\label{sec:introduction}
Large language model based agents are increasingly expected to
complete long-horizon tasks through sustained interaction with users, tools, and environments~\cite{hu2025hiagent, zhang2025survey}. These interactions produce growing histories of observations, actions, feedback, and task outcomes, making it inefficient and often infeasible to retain all past information within a single context. Effective agents must therefore preserve useful experience beyond the current context and retrieve it when relevant. Memory mechanisms provide this capability by organizing interaction histories into
persistent and reusable information that supports reasoning and
decision making across tasks and sessions~\cite{park2023generative, shinn2023reflexion, zhao2024expel, wang2023voyager, packer2023memgpt}. As agent trajectories become longer and more complex, determining what to store, how to organize it, and how to retrieve and update it has become a central problem in the design of long-term agent systems.

\begin{figure}
    \centering
    \includegraphics[width=0.9\linewidth]{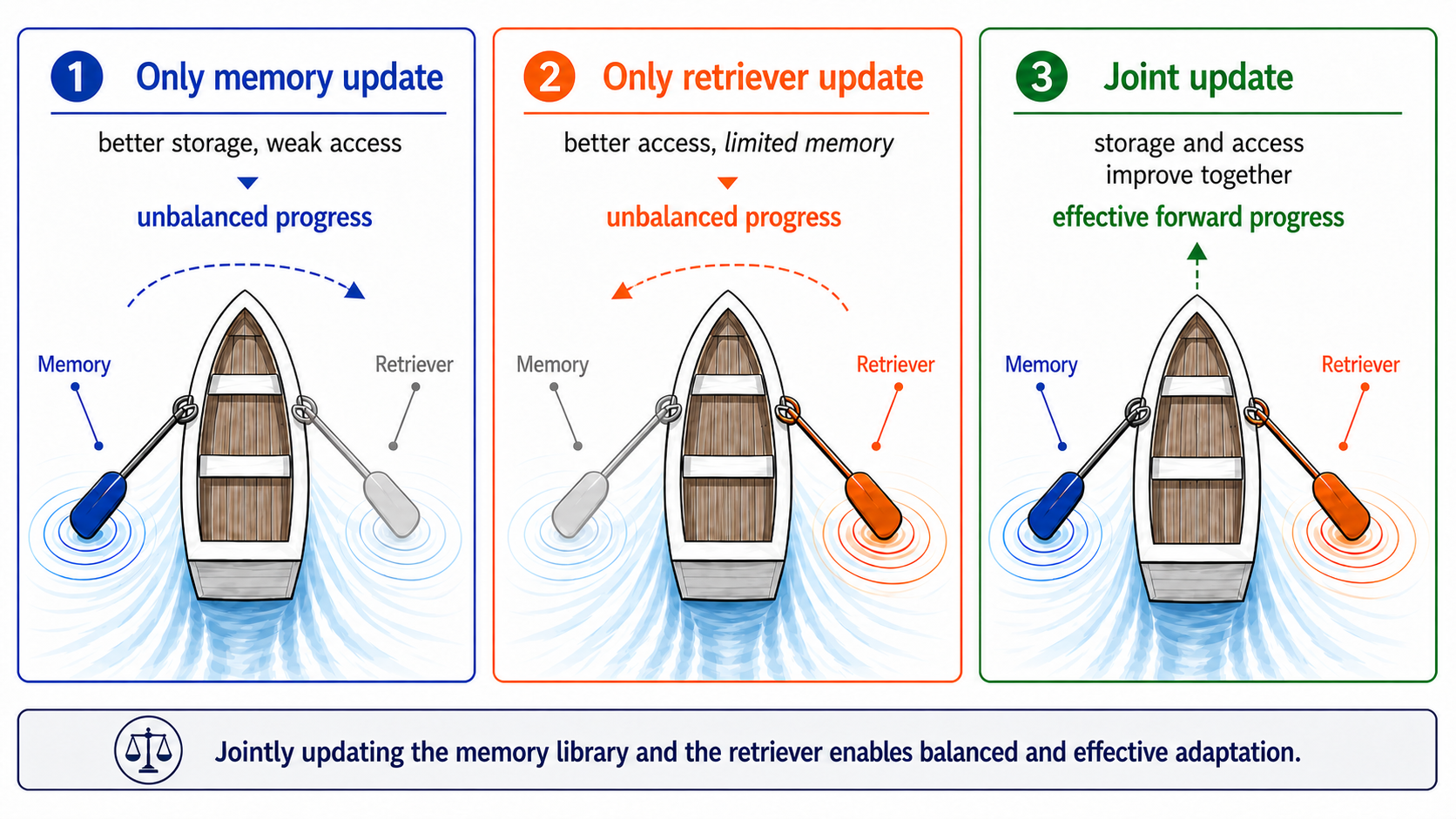}
    \caption{Three paradigms of Memory Systems.}
    \label{fig:moti}
\end{figure}

Existing work improves agent memory along two main directions. Memory-access methods use RAG, query reformulation, and adaptive retrieval to surface relevant experiences for each query~\cite{lewis2020retrieval, gao2023precise, asai2024self}. Memory-centric methods instead improve how experiences are updated, organized, or represented, ranging from value-based evolution such as MemQ and MemRL to structured stores and latent memory~\cite{liao2026memq, zhang2026memrl, xu2026mem, wang2024memoryllm}. These directions are often developed separately, with one component fixed while the other is optimized. Yet memory
access and evolution are mutually reinforcing: retrieval determines which memories are exposed and receive feedback, while memory updates reshape the values and organization that guide future retrieval.
As illustrated in Fig.~\ref{fig:moti}, updating the memory module or the retriever in isolation can improve memory quality or retrieval effectiveness, respectively, but prevents the fixed counterpart from adapting to the resulting changes. Joint adaptation instead allows the retrieval policy and memory bank to co-evolve within a unified feedback loop, thereby supporting sustained improvements in overall agent performance. This motivates us to formulate query routing and memory evolution as a coupled learning problem.

We propose \textbf{CoEvo-Mem}, a closed-loop framework that co-evolves retrieval policy and memory-bank refinement through interaction feedback. Rather than learning a retrieval policy from scratch, CoEvo-Mem uses a frozen LLM to generate channel-specific query rewrites and a confidence-weighted routing prior, while learning only a lightweight residual correction online. The corrected route weights control a value-aware hybrid retriever that integrates dense similarity, sparse matching, and learned memory utility. After generation, the interaction trajectory and retrieved memory set guide updates to both memory values and inter-memory relations, while response correctness provides the reward for adapting the residual router. To control the non-stationarity induced by this mutual dependence, CoEvo-Mem alternates between router adaptation and memory evolution.

We evaluate CoEvo-Mem on seven diverse benchmarks spanning long-term conversational memory, scientific and multimodal reasoning, code generation, operating-system interaction, and tool use. CoEvo-Mem achieves state-of-the-art performance across all benchmarks, empirically demonstrating the advantage of jointly optimizing memory access and memory evolution.

Our main contributions are threefold:
\begin{itemize}

\item We identify an underexplored feedback loop between retrieval and
memory evolution in long-term LLM agents. Retrieval determines both the
context available for generation and the memories that receive credit
from task outcomes; changes in memory utility estimates and relational
structure then influence subsequent retrieval.

\item We propose \textbf{CoEvo-Mem}, which coordinates retrieval policy
adaptation with relational memory evolution while keeping the answering
LLM frozen. It combines query rewrites for dense and sparse retrieval,
residual adaptation of an LLM routing prior from task rewards, and
memory updates that propagate feedback from interaction trajectories
across semantic, lexical, and temporal relations. Alternating phases
hold the inactive component fixed to limit nonstationarity.

\item Experiments on seven benchmarks show that CoEvo-Mem advances the
state of the art in every evaluated setting, with improvements of up to
\textbf{7.50 percentage points} over the strongest baseline. Ablation
studies further validate the complementary roles of retrieval
adaptation, relational memory evolution, and alternating training.

\end{itemize}

\section{Related Work}
\label{sec:related}

\subsection{Retrieval-Augmented Generation and Adaptive Access}

Retrieval-augmented generation (RAG)~\cite{lewis2020retrieval}
grounds LLM outputs in evidence retrieved at inference time. Its
retrieval layer has been improved through dense semantic
matching~\cite{karpukhin2020dense,khattab2020colbert,
santhanam2022colbertv2}, sparse lexical
matching~\cite{robertson2009probabilistic,formal2021splade}, and
rank-level fusion~\cite{cormack2009reciprocal}. Self-RAG learns when
to retrieve and critique evidence during generation~\cite{asai2024self},
while MiniRAG~\cite{fan2025minirag} and
LightRAG~\cite{guo2024lightrag} develop efficient retrieval pipelines
for accessing distributed evidence.

Adaptive retrieval further improves memory access at the query and
policy levels. HyDE~\cite{gao2023precise} generates hypothetical
documents for dense retrieval, and RAG-Fusion~\cite{rackauckas2024rag}
retrieves from multiple query variants. Search-R1~\cite{jin2025search}
optimizes multi-step search through reinforcement learning. These
methods improve evidence selection, but generally retrieve from a
fixed or independently maintained collection. CoEvo-Mem couples
adaptive retrieval routing with memory evolution, so that retrieval
decisions and memory updates can inform one another over time.

\subsection{Memory Representation and Evolution}

Agent memory is commonly implemented as either an editable external
store or a trainable model-integrated representation. Among external
systems, MemoryBank maintains personalized memories with a forgetting
mechanism~\cite{zhong2024memorybank}, while Mem0 extracts and
consolidates salient conversational facts~\cite{chhikara2025mem0}.
MemP represents reusable procedural experience~\cite{fang2026memp},
LangMem supports persistent memory extraction and
management~\cite{langchain2024langmem}, and MemoryOS organizes
long-term experience through a hierarchical memory
architecture~\cite{kang2025memory}. Structured approaches further
model dependencies among memories: A-MEM dynamically updates memory
attributes and links~\cite{xu2026mem}; Mem0$^{g}$, Zep, Memobase, and
StructMem maintain graph-based, temporal, or consolidated memory
structures~\cite{chhikara2025mem0,rasmussen2025zep,memobase2024,
xu2026structmem}.

Model-integrated methods learn memory representations or interfaces
within the neural computation process. LongMem accesses cached
representations through a trainable side network~\cite{wang2023augmenting},
MemoryLLM maintains a self-updatable latent memory
pool~\cite{wang2024memoryllm}, and AutoCompressor learns compact
context representations~\cite{chevalier2023adapting}. MemGen,
M+, and ElasticMem further explore generated, persistent, or
dynamically allocated latent memory~\cite{zhang2025memgen,
wang2025m+,feng2026elasticmem}. Meanwhile, MemRL and MemQ learn the
utility of external memories from task feedback and use it to guide
subsequent selection~\cite{zhang2026memrl,liao2026memq}. These methods
advance memory representation, organization, or valuation, while the
retrieval policy is typically predefined or optimized separately.
CoEvo-Mem instead co-adapts retrieval routing, memory values, and
inter-memory relations within a unified feedback loop.

\section{CoEvo-Mem Framework}
\label{sec:framework}

\begin{figure*}[t]
    \centering
    \includegraphics[
    width=1.0\linewidth,
    trim={0.3cm 0.0cm 0.1cm 0.0cm}, 
    clip]{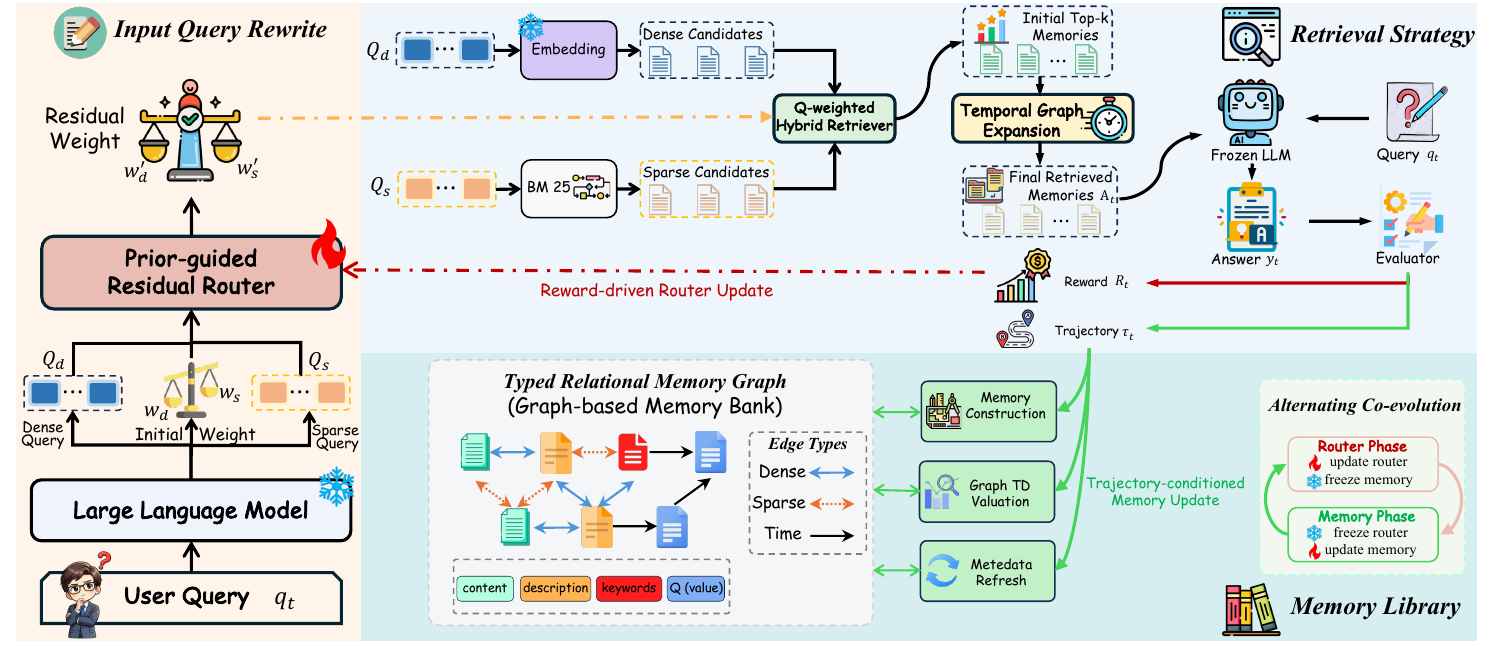}
    \caption{Overview of \textbf{CoEvo-Mem}. Route-specialized retrieval and relational memory evolution form a closed feedback loop and are co-adapted through alternating training phases.}
    \label{fig:framework}
\end{figure*}

\subsection{Coupled Retrieval--Memory Learning}
\label{subsec:overview}

Long-term memory creates a feedback loop that is absent from retrieval
over a static corpus. The retrieval policy determines both the context
supplied to the model and the memories eligible for task feedback,
while updates to memory utilities and relations reshape the retrieval
state encountered by subsequent queries. Optimizing either component
against a permanently fixed counterpart can therefore leave it
mismatched with the updated state of the other. CoEvo-Mem addresses this
coupled learning problem through phase-wise co-adaptation of retrieval
and relational memory, as summarized in
Figure~\ref{fig:framework}.

We consider a training stream
$\mathcal D_{\mathrm{tr}}=\{(q_i,\mathcal Y_i)\}_{i=1}^{T}$,
where $q_i$ is a query and $\mathcal Y_i$ specifies its task-dependent
evaluation target. Let
$r\in\{0,\ldots,N_{\mathrm{ph}}-1\}$ index training phases and $t$
index interactions within a phase. We write
$(q_{r,t},\mathcal Y_{r,t})$ for the $t$-th visited example and suppress
$r$ in interaction-level equations. CoEvo-Mem maintains a retrieval
policy $\pi_\theta$, implemented by the \emph{Prior-Guided Residual
Router}, and a persistent \emph{Typed Relational Memory Graph}
$\mathcal G_t=(\mathcal V_t,\mathcal E_t)$, while the answering LLM
$F$ remains frozen. Each node $m\in\mathcal V_t$ stores an experience
and an estimate $Q_t(m)$ of its discounted downstream retrieval utility.
Typed edges in $\mathcal E_t$ encode semantic, lexical, and temporal
relations aligned with the corresponding retrieval channels.

Given $q_t$, the frozen LLM produces route-specific rewrites and an
initial dense--sparse routing prior, which the residual router adapts
into fusion weights. Retrieval returns a bounded exposure set
\begin{equation}
\label{eq:forward-process}
\begin{aligned}
&(\widetilde{\mathcal Q}_t^d,
 \widetilde{\mathcal Q}_t^s,\mathbf w_t)
    =\operatorname{SRQR}_{F,\theta}(q_t),\\
&\mathcal A_t
    =\operatorname{Retrieve}
      (\widetilde{\mathcal Q}_t^d,
       \widetilde{\mathcal Q}_t^s,
       \mathcal G_t,\mathbf w_t),
\quad |\mathcal A_t|\leq k.
\end{aligned}
\end{equation}
The frozen LLM generates
$y_t\sim p_F(\cdot\mid q_t,\mathcal A_t)$, and a task-specific
evaluator returns $R_t=\operatorname{Eval}(y_t;\mathcal Y_t)\in[0,1]$.
The trajectory $\tau_t=(q_t,\mathcal A_t,y_t,R_t)$ drives either a
\emph{Reward-Driven Router Update} $\mathcal U_{\mathrm R}$ or a
\emph{Trajectory-Conditioned Memory Update} $\mathcal U_{\mathrm M}$,
with only one component active in a training phase. The retrieved set
$\mathcal A_t$ is their coupling interface: it
is induced by the router, provides context to $F$, and identifies the
memories eligible for outcome-conditioned credit.

We use expected task reward as a system-level training criterion:
\begin{equation}
\label{eq:sequence-objective}
J_{\mathrm{tr}}
=\mathbb E
\left[
\frac{1}{N_{\mathrm{ph}}T}
\sum_{r=0}^{N_{\mathrm{ph}}-1}
\sum_{t=1}^{T}R_{r,t}
\right].
\end{equation}
The expectation is over trajectories induced by the router, evolving
graph, and phase schedule. Rather than differentiating through the
discrete, stateful graph, CoEvo-Mem couples reward-adapted
route-specialized retrieval with retrieval-aligned relational memory,
alternating their updates while holding the other component fixed.

\subsection{Route-Specialized Query Rewriting and Retrieval}
\label{subsec:router}

Queries often contain both semantic intent and exact lexical cues, but
their relative importance varies by task. The \emph{Self-Routed Query
Rewriter} (SR-QR) separates these signals before retrieval. Under a
fixed routing prompt, $F$ returns
$\mathbf o_t=(\widetilde{\mathcal Q}_t^d,
\widetilde{\mathcal Q}_t^s,\mathbf p_t^0,\xi_t^0)
=F_{\mathrm{route}}(q_t)$. The dense rewrites preserve semantic intent,
the sparse rewrites emphasize entities and exact terms,
$\mathbf p_t^0\in\Delta^1$ is an initial dense--sparse routing prior,
and $\xi_t^0$ is an auxiliary confidence feature.

To adapt this prior from task outcomes, a trainable query encoder
$\phi_\psi$ and policy head $f_\omega$ produce the mean routing policy:
\begin{equation}
\label{eq:router-policy}
\begin{aligned}
\mathbf c_t
&=\phi_\psi(q_t,\mathbf o_t)\in\mathbb R^h,\\
\boldsymbol\Delta_t
&=f_\omega(\mathbf c_t)\in\mathbb R^2,\\
\bar p_t^0(a)
&=\frac{p_t^0(a)+\varepsilon}
{\sum_{a'\in\{d,s\}}\left(p_t^0(a')+\varepsilon\right)},
\quad a\in\{d,s\},\\
\boldsymbol\pi_t
&=\operatorname{softmax}\!\left(
\log\bar{\mathbf p}_t^0+\boldsymbol\Delta_t
\right).
\end{aligned}
\end{equation}
where $\theta=(\psi,\omega)$ and $\varepsilon>0$ smooths the prior.
The final layer of $f_\omega$ is
initialized to zero, so
$\boldsymbol\pi_t=\bar{\mathbf p}_t^0$ at initialization. Learning
therefore moves the policy toward deviations favored by downstream
reward rather than relearning routing from scratch.

During a router phase, the memory graph is fixed and CoEvo-Mem explores
continuous mixtures of the two routes. Specifically,
$\zeta_t\sim\operatorname{Beta}(\kappa\pi_t^d,\kappa\pi_t^s)$ with
$\kappa>0$, and retrieval uses
$\mathbf w_t=(\zeta_t,1-\zeta_t)$. The conditional mean of
$\mathbf w_t$ is $\boldsymbol\pi_t$. Let
$B_t=(1-\lambda_{\mathrm{ema}})B_{t-1}
+\lambda_{\mathrm{ema}}R_t$ be an exponential moving-average baseline
and $\widehat A_t=R_t-B_{t-1}$. Following the score-function estimator,
the router minimizes
\begin{equation}
\label{eq:router-loss}
\begin{aligned}
\widehat{\mathcal L}_{\mathrm{router},t}(\theta)
={}&-\operatorname{sg}[\widehat A_t]
 \log p_\theta(\zeta_t\mid q_t,\mathbf o_t)\\
&+\beta_{\mathrm{KL}}D_{\mathrm{KL}}\!\left(
\operatorname{Cat}(\boldsymbol\pi_t)
\,\middle\|\,
\operatorname{Cat}(\bar{\mathbf p}_t^0)
\right).
\end{aligned}
\end{equation}
Here, $p_\theta$ is the Beta density above,
$\operatorname{sg}$ denotes stop-gradient, and $\beta_{\mathrm{KL}}$
controls prior regularization. The score-function
term~\cite{williams1992simple,ahmadian2024back} assigns terminal task
reward through the non-differentiable retrieval process; the second
discourages large departures from the LLM prior. Exploration is disabled during memory
phases and inference, where retrieval uses
$\mathbf w_t=\boldsymbol\pi_t$.

For each rewrite, dense retrieval ranks memories by description
embedding similarity, while sparse retrieval ranks them by BM25 over
entities and keywords~\cite{robertson2009probabilistic}. Let
$\mathcal L_{t,a,i}$ be the ranked list returned by rewrite $i$ on
route $a\in\{d,s\}$, with $N_a$ valid rewrites on that route; their union forms the initial pool
$\mathcal P_t^{(0)}$.
For any candidate pool $\mathcal P$, each rewrite score and $Q_t$ induce
a rank over all candidates in $\mathcal P$. The \emph{$Q$-Weighted
Hybrid Retriever} then fuses route relevance with learned memory utility
using weighted reciprocal rank fusion~\cite{cormack2009reciprocal}:
\begin{equation}
\label{eq:hybrid-score}
\begin{aligned}
S_t(m;\mathcal P)
={}&
\sum_{a\in\{d,s\}}
\frac{w_t^a}{N_a}
\sum_{i=1}^{N_a}
\frac{1}
{\eta+\operatorname{rank}_{a,i}^t(m;\mathcal P)}\\
&+
\frac{\lambda_Q}
{\eta+\operatorname{rank}_{Q}^t(m;\mathcal P)}.
\end{aligned}
\end{equation}
Here, $\eta>0$ is the RRF constant and $\lambda_Q\geq0$ controls the
utility rank. The $1/N_a$ factor prevents a route with more rewrites
from receiving more mass by construction. Let $\mathcal K_t$ be the top-$k$ candidates
under $S_t(\cdot;\mathcal P_t^{(0)})$. The \emph{Temporal Graph
Expansion} step augments the initial pool with the incoming and outgoing
\textsc{Time} neighbors of $\mathcal K_t$, recomputes all route-specific
and utility ranks on the expanded pool, and returns its top $k$ memories.
Thus, unlike a fixed ensemble,
route-specialized rewrites define the evidence lists, an outcome-adapted
mixture controls their exposure, and evolving utilities and temporal
relations alter subsequent rankings.

\subsection{Retrieval-Aligned Relational Memory Evolution}
\label{subsec:memory_evolution}

The memory graph mirrors the evidence channels used by retrieval. A node
stores preserved content $x_m$, a dense description $d_m$, sparse
keywords $K_m$, temporal metadata $s_m$, source trajectory $\tau_m$,
and utility $Q_t(m)$. The relation vocabulary is
$\mathcal C=\{\textsc{Dense},\textsc{Sparse},\textsc{Time}\}$.
Dense and sparse edges are reciprocal and encode semantic proximity and
lexical overlap, respectively; a time edge is directed from an earlier
memory to a later continuation. Hence, the fields and relations of
$\mathcal G_t$ directly match the dense, sparse, and temporal operations
used by retrieval, rather than forming an auxiliary generic knowledge
graph.

In a memory phase, the \emph{Memory Construction} step uses a fixed
prompt to distill every training trajectory, regardless of outcome,
into a new memory $m_t^{\mathrm{new}}$. The reward is recorded as
provenance rather than used for filtering. Candidate dense and sparse
edges are derived from embedding similarity and keyword overlap, while
temporal links are verified by the frozen LLM and oriented by timestamp.
Both construction and valuation observe only the generated trajectory
and scalar task outcome, not the gold target, a corrected answer, or
evaluator rationale.

During \emph{Graph TD Valuation}, a fixed utility prompt assesses each
exposed memory $m\in\mathcal A_t$ using the query, complete generated
trajectory, and scalar outcome, and returns a contribution score
$u_t(m)\in[0,1]$. Because a new memory has no reuse history, its initial utility is inherited from the local context that produced it:
\begin{equation}
\label{eq:new-memory-value}
\widehat Q_t(m_t^{\mathrm{new}})
=
\begin{cases}
|\mathcal A_t|^{-1}
\sum_{m\in\mathcal A_t}Q_t(m),
&\mathcal A_t\neq\varnothing,\\
Q_{\mathrm{init}},&\mathcal A_t=\varnothing,
\end{cases}.
\end{equation}
$Q_{\mathrm{init}}$ is a task-independent neutral initialization.
We treat the memory distilled from the current interaction as the
successor experience for bootstrapping. For each exposed memory, the
resulting TD-style utility residual is
\begin{equation}
\label{eq:memory-td}
\delta_t(m)
=R_t\,u_t(m)+\gamma\widehat Q_t(m_t^{\mathrm{new}})-Q_t(m),
\quad m\in\mathcal A_t,
\end{equation}
where $\gamma\in[0,1)$ is the discount factor. The immediate term
combines task outcome with memory-specific attribution, while the
bootstrap term transfers downstream utility from the newly distilled
experience to the context that supported its construction.

Credit also propagates to related memories. For a path $p:v\rightsquigarrow
m$ of at most $D$ edges, let its strength be the product of
relation-specific attenuation factors $\rho_c\in[0,1]$. We retain the
strongest valid path,
\begin{equation}
\label{eq:relational-credit}
c_t^{(D)}(v,m)
=
\max_{\substack{p:v\rightsquigarrow m\\1\leq |p|\leq D}}
\prod_{e\in p}\rho_{c(e)}.
\end{equation}
We set $c_t^{(D)}(v,m)=0$ when no valid path exists. Taking the maximum
prevents path multiplicity from mechanically amplifying credit. Let
$\mathcal S_t(v)=
\{m\in\mathcal A_t:c_t^{(D)}(v,m)>0\}$. For
$\mathcal S_t(v)\neq\varnothing$, define
\begin{equation}
\label{eq:propagated-td}
\begin{aligned}
\bar\delta_t(v)
&=\frac{1}{|\mathcal S_t(v)|}
\sum_{m\in\mathcal S_t(v)}c_t^{(D)}(v,m)\delta_t(m),\\
\Delta_t(v)
&=
\begin{cases}
\delta_t(v),&v\in\mathcal A_t,\\
\bar\delta_t(v),& v\notin\mathcal A_t,\mathcal S_t(v)\neq\varnothing\\
0,&\text{otherwise}.
\end{cases}
\end{aligned}
\end{equation}
The averaging over $\mathcal S_t(v)$ controls update scale when a memory
is connected to multiple exposed items. Dense and sparse edges can be
traversed in both directions. Time edges are traversed only from earlier
to later memories; thus, along a purely temporal path
$v\rightsquigarrow m$, credit from an exposed memory $m$ is assigned to
its chronological predecessors.

All residuals for nodes present before insertion are computed from the
same pre-update snapshot and applied synchronously:
\begin{equation}
\label{eq:memory-value-update}
Q_{t+1}(v)
=
\Pi_{[Q_{\min},Q_{\max}]}
\bigl(Q_t(v)+\alpha_Q\Delta_t(v)\bigr),
\quad v\in\mathcal V_t.
\end{equation}
Here, $\alpha_Q>0$ is the update rate and $\Pi$ projects utility onto
the admissible interval $[Q_{\min},Q_{\max}]$.
The new node is assigned $\widehat Q_t(m_t^{\mathrm{new}})$ and becomes
eligible for value updates from the next interaction. Feedback is thus
not confined to the items returned by one query; it also refines related
regions while controlling propagation by relation type and path depth.

\subsection{Alternating Co-Evolution}
\label{subsec:alternating}

Simultaneously updating the router and memory graph couples two moving
processes: routing changes the memory-exposure distribution, while
memory evolution changes the retrieval landscape. CoEvo-Mem therefore
adopts a phase-wise block update schedule in which only one component
is active at a time. Each phase makes one complete pass over
$\mathcal D_{\mathrm{tr}}$, and repeated alternation exposes each
component to the latest state of the other without updating both from
the same trajectory.

Let $z_r\in\{\textsc{Router},\textsc{Memory}\}$ denote the active phase.
We define the corresponding full-pass updates as
\[
\theta_{\mathrm R}^{(r+1)}
=
\operatorname{Pass}_{\mathrm R}
(\theta^{(r)};\mathcal G^{(r)},\mathcal D_{\mathrm{tr}})
\]
and
\[
\mathcal G_{\mathrm M}^{(r+1)}
=
\operatorname{Pass}_{\mathrm M}
(\mathcal G^{(r)};\theta^{(r)},\mathcal D_{\mathrm{tr}}).
\]
The phase-level state transition is
\begin{equation}
\label{eq:alternating-update}
(\theta^{(r+1)},\mathcal G^{(r+1)})
=
\begin{cases}
(\theta_{\mathrm R}^{(r+1)},\mathcal G^{(r)}),
&z_r=\textsc{Router},\\
(\theta^{(r)},\mathcal G_{\mathrm M}^{(r+1)}),
&z_r=\textsc{Memory}.
\end{cases}
\end{equation}
A router pass keeps the graph read-only throughout the phase. A memory
pass instead keeps the router fixed while the graph evolves
sequentially over the training interactions.

Equation~\eqref{eq:alternating-update} defines the phase-level block
update. Each phase makes exactly one complete pass over
$\mathcal D_{\mathrm{tr}}$. When task histories permit graph
initialization, training begins with a \textsc{Router} phase and then
alternates between \textsc{Router} and \textsc{Memory} phases. If the
graph is initially empty, a single \textsc{Memory} phase is prepended
to bootstrap it before the same alternating schedule is applied.

Both phases follow the same interaction pipeline but differ in the
component being optimized. A \textsc{Router} phase keeps the graph
read-only, samples retrieval mixtures, and updates only $\theta$. A
\textsc{Memory} phase fixes the router, uses its deterministic mean
policy, and updates $\mathcal G$ sequentially as training examples are
processed. Thus, each component is optimized against the latest state
of its counterpart while cross-component updates remain separated
within a phase.

All router and graph updates are confined to the training split.
During validation, both components remain fixed, and validation
performance is used solely for checkpoint selection. The selected
router and memory state are then frozen and evaluated once on the test
split using deterministic retrieval, without consuming test outcomes
or modifying the memory graph. Algorithm~\ref{alg:coevo-training}
summarizes the resulting training procedure.
\begin{algorithm}[t]
\small
\caption{Phase-wise Training of CoEvo-Mem}
\label{alg:coevo-training}
\begin{algorithmic}[1]

\Statex \textbf{Input:} Training set $\mathcal D_{\mathrm{tr}}$,
frozen LLM $F$, initialized router $\theta$, memory graph
$\mathcal G=(\mathcal V,\mathcal E)$, and phase schedule $\mathsf S$
\Statex \textbf{Output:} Trained router $\theta^\star$ and memory graph
$\mathcal G^\star$

\If{$\mathcal V=\varnothing$}
    \State Prepend \textsc{Memory} to $\mathsf S$
\EndIf

\For{$z_r\in\mathsf S$}
    \ForAll{$(q_t,\mathcal Y_t)\in\mathcal D_{\mathrm{tr}}$}
        \State $\mathbf o_t\gets F_{\mathrm{route}}(q_t)$;
        $\boldsymbol\pi_t\gets\pi_\theta(q_t,\mathbf o_t)$

        \If{$z_r=\textsc{Router}$}
            \State $\zeta_t\sim
            \operatorname{Beta}(\kappa\pi_t^d,\kappa\pi_t^s)$
            \State $\mathbf w_t\gets(\zeta_t,1-\zeta_t)$
        \Else
            \State $\mathbf w_t\gets\boldsymbol\pi_t$
        \EndIf

        \State $\mathcal A_t\gets\operatorname{Retrieve}
        (\widetilde{\mathcal Q}_t^d,
         \widetilde{\mathcal Q}_t^s,
         \mathcal G,\mathbf w_t)$
        \State $y_t\sim p_F(\cdot\mid q_t,\mathcal A_t)$;
        $R_t\gets\operatorname{Eval}(y_t;\mathcal Y_t)$
        \State $\tau_t\gets(q_t,\mathcal A_t,y_t,R_t)$

        \If{$z_r=\textsc{Router}$}
            \State $\theta\gets
            \mathcal U_{\mathrm R}
            (\theta;q_t,\mathbf o_t,\zeta_t,R_t)$
        \Else
            \State $\mathcal G\gets
            \mathcal U_{\mathrm M}
            (\mathcal G;\mathcal A_t,\tau_t,F)$
        \EndIf
    \EndFor
\EndFor

\State $\theta^\star\gets\theta$;
$\mathcal G^\star\gets\mathcal G$
\State \Return $(\theta^\star,\mathcal G^\star)$

\end{algorithmic}
\end{algorithm}


\begin{table*}[!t]
\centering
\begin{threeparttable}

\scriptsize
\setlength{\tabcolsep}{2.8pt}
\renewcommand{\arraystretch}{1.08}

\begin{tabularx}{\textwidth}{@{}l|l|*{8}{>{\centering\arraybackslash}X}@{}}
\toprule
\textbf{Benchmark} 
& \textbf{Model} 
& \makecell{\textbf{No} \textbf{Mem.}} 
& \textbf{RAG} 
& \makecell{\textbf{Self-}\textbf{RAG}} 
& \textbf{Mem0} 
& \textbf{MemP} 
& \textbf{MemRL} 
& \textit{\textbf{MemQ}} 
& \textbf{Ours} \\
\midrule

\makecell[l]{LLAB\\[-1pt]\textit{os interaction}}
& 4o-mini
& \(66.89_{\pm 1.26}\)
& \(68.89_{\pm 0.38}\)
& \(70.45_{\pm 0.39}\)
& \(70.00_{\pm 0.67}\)
& \(71.11_{\pm 0.38}\)
& \(74.44_{\pm 1.02}\)
& \(\underline{74.67_{\pm 0.67}}\)
& \cellcolor{ourspeach}\(\mathbf{76.00}_{\pm 0.82}\) \\

\midrule

\makecell[l]{LiveCodeBench\\[-1pt]\textit{coding}}
& \multirow{3}{*}{\makecell{Gemma-\\4-E4B-it}}
& \(44.76_{\pm 2.29}\)
& \(50.48_{\pm 1.65}\)
& \(49.52_{\pm 1.65}\)
& \(47.62_{\pm 1.65}\)
& \(49.52_{\pm 1.65}\)
& \(45.71_{\pm 2.86}\)
& \(\underline{51.43_{\pm 2.86}}\)
& \cellcolor{ourspeach}\(\mathbf{55.24}_{\pm 1.34}\) \\

\makecell[l]{MMMU Pro\\[-1pt]\textit{multimodal}}
&
& \(48.46_{\pm 0.76}\)
& \(53.76_{\pm 1.04}\)
& \(53.57_{\pm 0.84}\)
& \(49.90_{\pm 1.37}\)
& \(54.24_{\pm 1.42}\)
& \(53.66_{\pm 1.01}\)
& \(\underline{54.43_{\pm 0.88}}\)
& \cellcolor{ourspeach}\(\mathbf{55.49}_{\pm 1.27}\) \\

\makecell[l]{GPQA Diamond\\[-1pt]\textit{science qa}}
&
& \(47.50_{\pm 3.01}\)
& \(58.33_{\pm 3.82}\)
& \(58.33_{\pm 3.82}\)
& \(58.33_{\pm 1.18}\)
& \(59.17_{\pm 1.44}\)
& \(\underline{60.83_{\pm 5.20}}\)
& \(\underline{60.83_{\pm 3.82}}\)
& \cellcolor{ourspeach}\(\mathbf{68.33}_{\pm 3.11}\) \\

\midrule

\makecell[l]{BFCL\\[-1pt]\textit{function call}}
& \makecell{Qwen3.5-\\35B-A3B$^{\ast}$}
& \(33.67_{\pm 3.09}\)
& \(35.33_{\pm 2.49}\)
& \(38.00_{\pm 2.83}\)
& \(35.00_{\pm 2.45}\)
& \(37.33_{\pm 1.70}\)
& \(\underline{39.00_{\pm 0.00}}\)
& \(37.33_{\pm 2.08}\)
& \cellcolor{ourspeach}\(\mathbf{45.33}_{\pm 2.05}\) \\

\midrule

\makecell[l]{BFCL\\[-1pt]\textit{function call}}
& \makecell{Qwen3.5-\\35B-A3B$^{\ddagger}$}
& \(56.71_{\pm 1.71}\)
& \(55.45_{\pm 2.62}\)
& \(60.07_{\pm 2.49}\)
& \(54.79_{\pm 1.68}\)
& \(61.39_{\pm 0.99}\)
& \(60.07_{\pm 2.29}\)
& \(\underline{62.38_{\pm 1.71}}\)
& \cellcolor{ourspeach}\(\mathbf{64.69}_{\pm 0.93}\) \\

\bottomrule
\end{tabularx}
\caption{Performance comparison across benchmarks. Results are reported as mean $\pm$ standard deviation.
The best and second-best mean scores are highlighted in bold and
underlined, respectively. $^{\ast}$ and $^{\ddagger}$ denote the
non-thinking and thinking settings.}\label{tab:main_results}

\end{threeparttable}
\end{table*}

\section{Experiments}

\subsection{Experimental Settings}
\paragraph{Benchmarks.}
We evaluate CoEvo-Mem on seven benchmarks covering operating-system
interaction (LLAB~\cite{zheng2025lifelongagentbench}), code generation
(LiveCodeBench~\cite{jain2025livecodebench}), multimodal reasoning
(MMMU Pro~\cite{yue-etal-2025-mmmu}), scientific question answering (GPQA
Diamond~\cite{rein2023gpqa}), function calling
(BFCL~\cite{patil2025berkeley}), and long-term memory
(LoCoMo~\cite{maharana2024evaluating} and
LongMemEval~\cite{wu2024longmemeval}). For LoCoMo and LongMemEval, we
follow the MC-10 protocol of ElasticMem~\cite{feng2026elasticmem},
which casts each query as a 10-way multiple-choice task. We also
evaluate LoCoMo under its original free-form QA setting to assess
open-ended conversational memory. All dataset splits are fixed before optimization, with router and memory updates restricted to the training set and both components frozen during test.
\paragraph{Baselines.}
We compare CoEvo-Mem with representative retrieval-augmented,
external-memory, structured-memory, and trainable memory methods.
Baselines are selected according to task and backbone compatibility,
with \emph{No Memory} and \emph{Full Context} included as reference configurations.

\paragraph{Model configurations.}
We follow the established answering and embedding configurations for
each benchmark whenever available. Our primary embedding models are
Qwen3-Embedding-8B and text-embedding-3-small. For LoCoMo free-form
QA, we use GPT-4o-mini as the answering backbone,
text-embedding-3-small for retrieval, and DeepSeek-V4-Pro as the
response evaluator.


\subsection{Long-Term Memory Evaluation on Diverse Agent Tasks}

As shown in Table~\ref{tab:main_results}, CoEvo-Mem achieves the
highest mean score in all six evaluated settings. Compared with the
strongest baseline in each row, the average improvement is $3.72$
percentage points, with individual gains ranging from $1.06$ to $7.50$
points. The largest improvements occur on GPQA Diamond ($+7.50$) and
BFCL in the non-thinking setting ($+6.33$), followed by LiveCodeBench
($+3.81$). Notably, these three settings span scientific question
answering, tool use, and code generation, suggesting that the gains are
not concentrated in a single task domain or experience format.
Positive improvements on LLAB, MMMU Pro, and BFCL with thinking further
show that the advantage persists across interaction, multimodal, and
different inference settings. Overall, the consistent gains across
heterogeneous retrieval demands and memory types support the generality
of co-adapting retrieval routing and memory evolution.

\begin{figure}[t]
    \centering
    \includegraphics[width=0.96\linewidth]{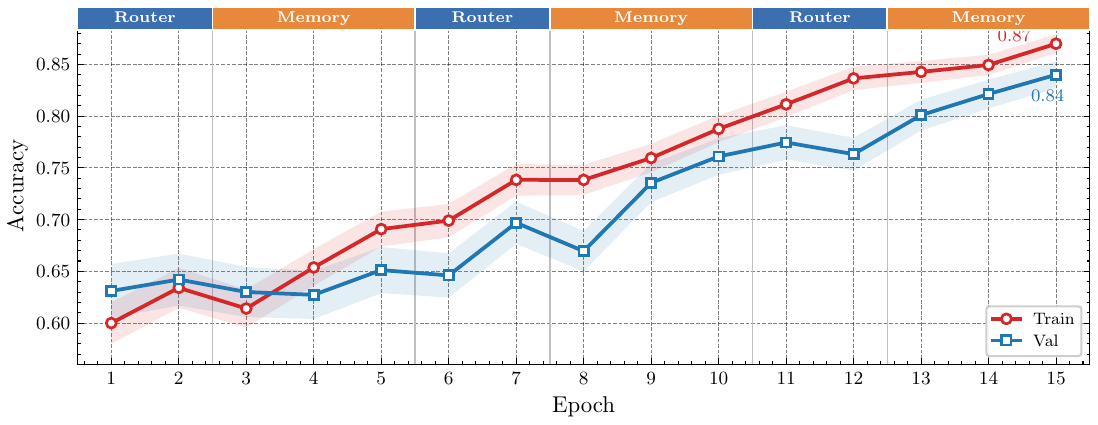}
    \caption{Training success rate on LoCoMo free-form QA.}
    \label{fig:locomo_training}
\end{figure}

\begin{table}[t]
\centering
\begin{threeparttable}

\scriptsize
\setlength{\tabcolsep}{3pt}
\renewcommand{\arraystretch}{1.08}

\begin{tabularx}{\columnwidth}
{@{}l|>{\centering\arraybackslash}X
@{\vDashSep}
*{4}{>{\centering\arraybackslash}X}@{}}
\toprule
\textbf{Method}
& \textbf{Overall} $\uparrow$
& \textbf{Multi} $\uparrow$
& \textbf{Open} $\uparrow$
& \textbf{Single} $\uparrow$
& \textbf{Temp} $\uparrow$ \\
\midrule

\rowcolor{raggreen}
OpenAI
& 71.82
& 69.86
& 53.12
& 84.66
& 45.48 \\

\rowcolor{raggreen}
FullContext
& 73.83
& 68.79
& 56.25
& \textbf{86.56}
& 50.16 \\


\rowcolor{raggreen}
LightRAG
& 68.83
& 66.31
& 50.00
& 77.53
& 53.89 \\

\midrule

\rowcolor{flatblue}
LangMem
& 58.10
& 62.23
& 47.92
& 71.12
& 23.43 \\

\rowcolor{flatblue}
A-Mem
& 64.16
& 56.03
& 31.25
& 72.06
& 60.44 \\

\rowcolor{flatblue}
Mem0
& 66.88
& 67.13
& 51.15
& 72.93
& 59.19 \\

\midrule

\rowcolor{structpink}
MemoryOS
& 58.25
& 56.74
& 45.83
& 67.06
& 40.19 \\


\rowcolor{structpink}
Zep
& 75.14
& \textbf{74.11}
& \underline{66.04}
& 79.79
& 67.71 \\

\rowcolor{structpink}
Memobase
& 75.78
& 70.92
& 46.88
& 77.17
& \textbf{85.05} \\

\rowcolor{structpink}
StructMem
& \underline{76.82}
& 68.77
& 46.88
& 81.09
& 81.62 \\

\midrule

\rowcolor{ourspeach}
Ours
& \textbf{81.71}
& \underline{72.22}
& \textbf{71.42}
& \underline{86.44}
& \underline{83.87} \\

\bottomrule
\end{tabularx}

\caption{
Performance on LoCoMo QA task.
{\setlength{\fboxsep}{1.5pt}\colorbox{raggreen}{Green}},
{\setlength{\fboxsep}{1.5pt}\colorbox{flatblue}{Blue}}, and
{\setlength{\fboxsep}{1.5pt}\colorbox{structpink}{Pink}}
denote context and retrieval baselines, external-memory methods,
and structured-memory methods, respectively.
}
\label{tab:locomo_single_column}

\end{threeparttable}
\end{table}

\begin{table*}[t]
\centering
\begin{threeparttable}

\scriptsize
\setlength{\tabcolsep}{4pt}
\renewcommand{\arraystretch}{1.08}

\begin{tabular*}{\textwidth}
{@{\extracolsep{\fill}}l|ccccc@{}}
\toprule
\textbf{Benchmark}
& \textbf{Full}
& \textbf{w/o SR-QR}
& \makecell{\textbf{w/o Memory} \textbf{Evolution}}
& \textbf{Simultaneous}
& \textbf{Two-Stage} \\
\midrule

GPQA Diamond
& \(\mathbf{68.33}\)
& \(60.00\)
& \(58.33\)
& \(66.67\)
& \(63.33\)\\

LiveCodeBench
& \(\mathbf{55.24}\)
& \(50.48\)
& \(47.62\)
& \(52.38\)
& \(51.43\)\\

LLAB
& \(\mathbf{76.00}\)
& \(74.00\)
& \(74.00\)
& \(75.00\)
& \(74.00\)\\

MMMU Pro
& \(\mathbf{55.49}\)
& \(52.60\)
& \(53.28\)
& \(53.95\)
& \(52.60\)\\

BFCL, no-think
& \(\mathbf{45.33}\)
& \(43.00\)
& \(43.33\)
& \(44.33\)
& \(43.33\)\\

BFCL, thinking
& \(\mathbf{64.69}\)
& \(59.74\)
& \(58.75\)
& \(61.39\)
& \(60.06\)\\

\midrule

LoCoMo-MC$^{*}$
& \(\mathbf{71}\)
& \(65\)
& \textemdash
& \textemdash
& \textemdash \\

LongMemEval-MC$^{*}$
& \(\mathbf{64}\)
& \(54\)
& \textemdash
& \textemdash
& \textemdash \\

\midrule

LoCoMo
& \(\mathbf{81.71}\)
& \(77.29\)
& \(77.88\)
& \(79.65\)
& \(78.76\)\\

\bottomrule
\end{tabular*}
\caption{Ablation study of CoEvo-Mem across different benchmarks. Results from repeated runs are reported as mean. $^{*}$ denotes Qwen2.5-7B-Instruct. Memory-evolution and schedule ablations are not applicable to the two fixed-memory MC protocols.}\label{tab:ablation_study}

\end{threeparttable}
\end{table*}

\begin{table}[t]
\centering
\begin{threeparttable}

\scriptsize
\setlength{\tabcolsep}{3pt}
\renewcommand{\arraystretch}{1.08}

\begin{tabularx}{\columnwidth}
{@{}c|l|>{\centering\arraybackslash}X
>{\centering\arraybackslash}X@{}}
\toprule
\textbf{Model}
& \textbf{Method}
& \makecell{\textbf{LoCoMo-MC}}
& \makecell{\textbf{LongMemEval-MC}} \\
\midrule

\multirow{8}{*}{
\rotatebox{90}{\textbf{Qwen2.5-3B-Instruct}}}
& A-MEM      & 39 & 45 \\
& Mem0       & 33 & 33 \\
& MemoryOS   & 33 & 40 \\
& MemP       & 26 & 41 \\
& LangMem    & 40 & 28 \\

& \cellcolor{trainpurple}MemGen
& \cellcolor{trainpurple}60
& \cellcolor{trainpurple}55 \\

& \cellcolor{trainpurple}AutoCompressor
& \cellcolor{trainpurple}47
& \cellcolor{trainpurple}44 \\


& \cellcolor{trainpurple}ElasticMem
& \cellcolor{trainpurple}\textbf{74}
& \cellcolor{trainpurple}\textbf{68} \\

& \cellcolor{ourspeach}Ours
& \cellcolor{ourspeach}\underline{63}
& \cellcolor{ourspeach}\underline{56} \\

\midrule

\multirow{8}{*}{
\rotatebox{90}{\textbf{Qwen2.5-7B-Instruct}}}
& A-MEM      & 42 & 50 \\
& Mem0       & 38 & 48 \\
& MemoryOS   & 42 & 53 \\
& MemP       & 22 & 48 \\
& LangMem    & 35 & 42 \\

& \cellcolor{trainpurple}MemGen
& \cellcolor{trainpurple}68
& \cellcolor{trainpurple}63 \\

& \cellcolor{trainpurple}AutoCompressor
& \cellcolor{trainpurple}55
& \cellcolor{trainpurple}51 \\


& \cellcolor{trainpurple}ElasticMem
& \cellcolor{trainpurple}\textbf{84}
& \cellcolor{trainpurple}\textbf{76} \\

& \cellcolor{ourspeach}Ours
& \cellcolor{ourspeach}\underline{71}
& \cellcolor{ourspeach}\underline{64} \\

\midrule

\textbf{GPT-4}
& \cellcolor{ourspeach}Ours
& \cellcolor{ourspeach}\textcolor{red}{89}
& \cellcolor{ourspeach}\textcolor{red}{82} \\

\bottomrule
\end{tabularx}
\caption{Multiple-choice accuracy on LoCoMo-MC and LongMemEval-MC
under fixed-memory evaluation. {\setlength{\fboxsep}{1.5pt}\colorbox{trainpurple}{Purple}} marks
trainable model-integrated methods, and
{\setlength{\fboxsep}{1.5pt}\colorbox{ourspeach}{Peach}} marks
CoEvo-Mem. GPT-4 is a stronger-backbone reference.}
\label{tab:selected_memory_qa}


\end{threeparttable}
\end{table}

\subsection{Long-Term Conversational Memory Task}

We evaluate two complementary aspects of long-term conversational
memory. LoCoMo free-form QA evaluates the complete CoEvo-Mem loop,
including router adaptation and memory evolution. QA instances are
divided into training, validation, and test sets over five independent
splits, and generated answers are evaluated by DeepSeek-V4-Pro.
Table~\ref{tab:locomo_single_column} shows that CoEvo-Mem achieves the
highest overall score of $81.71$, exceeding StructMem by $4.89$
percentage points and ranking first or second in every question
category. This consistency indicates that the improvement is not driven
by a single type of conversational-memory reasoning.
Figure~\ref{fig:locomo_training} further shows that training and
validation accuracy increase to $87\%$ and $84\%$, respectively,
suggesting that the gains from alternating co-adaptation generalize to
held-out questions.

LoCoMo-MC and LongMemEval-MC isolate the retrieval component through
conversation-level splits. The router is trained on graphs from the
training conversations and evaluated on fixed graphs constructed from
disjoint test conversations; memory evolution is disabled in this
protocol. CoEvo-Mem ranks second in all matched-backbone comparisons,
achieving $63\%/56\%$ and $71\%/64\%$ with the 3B and 7B backbones,
respectively, and performs best among methods that keep
model-integrated components frozen. Using GPT-4 as the frozen answering
backbone raises the scores to $89\%/82\%$ without retraining the memory
system, demonstrating that the learned retrieval interface transfers
across answering backbones.

\subsection{Ablation Study}

We conduct four controlled ablations to evaluate the two core
components of CoEvo-Mem and their optimization schedule.
\textbf{w/o SR-QR} removes route-specific query rewriting and
prior-guided residual routing, using the original query with fixed
dense and sparse fusion weights. \textbf{w/o Memory Evolution}
represents memories as independent items and retains only direct
item-level updates, removing typed relations and graph-based value
propagation. \textbf{Simultaneous} updates both components within each
training round, while \textbf{Two-Stage} optimizes them in two
consecutive phases. The schedule variants use the same task order and
component-update budget.

Table~\ref{tab:ablation_study} shows that removing SR-QR and memory
evolution reduces the macro-average score by $4.24$ and $4.80$
percentage points, respectively. SR-QR produces substantial gains on
GPQA Diamond and reduces the error on LoCoMo-MC and LongMemEval-MC by
$6.00$ and $10.00$ points. Since the MC protocols use fixed memory
graphs, these improvements directly reflect better query formulation
and retrieval routing. Memory evolution contributes most strongly to
GPQA Diamond, LiveCodeBench, and BFCL in the thinking setting, where
its removal causes drops of $10.00$, $7.62$, and $5.94$ points. This
confirms that typed relations and value propagation improve memory
reuse beyond independent item-level updates.

The schedule comparison further isolates how the two components should
be coordinated. Simultaneous training is consistently the second-best
variant, while two-stage training produces a larger degradation.
Alternating optimization outperforms them by $1.92$ and $3.33$
percentage points on average, respectively. Updating each component
against a fixed counterpart within a phase, followed by repeated
switching, provides more effective coordination between retrieval
routing and memory evolution.

\section{Conclusion}

We presented \textbf{CoEvo-Mem}, a framework that co-adapts retrieval routing and relational memory while keeping the base LLM frozen. Its SR-QR module rewrites queries for dense and sparse retrieval and learns a reward-adapted correction to the LLM routing prior. Its retrieval-aligned memory graph organizes experiences through semantic, lexical, and temporal relations, enabling outcome-conditioned utility updates and relational credit propagation. These components are coordinated through alternating full-pass phases that update one component while holding the other fixed.

Across seven benchmarks, CoEvo-Mem achieves state-of-the-art
performance in the reported settings. Comparisons and ablations
demonstrate the contributions of route-specialized retrieval,
relational memory evolution, and phase-wise co-adaptation across
heterogeneous tasks and memory types. These results highlight the
importance of jointly designing how experiences are retrieved,
evaluated, and evolved in long-term memory systems. Future work will explore multi-agent memory sharing and theoretical guarantees for coupled retrieval--memory optimization.

\bibliography{aaai2027}



\clearpage
%
%
%

\providecommand{\method}{\textsc{CoEvo-Mem}}
\providecommand{\Dense}{\textsc{Dense}}
\providecommand{\Sparse}{\textsc{Sparse}}
\providecommand{\TimeRel}{\textsc{Time}}
\providecommand{\RouterPhase}{\textsc{Router}}
\providecommand{\MemoryPhase}{\textsc{Memory}}
\providecommand{\JointPhase}{\textsc{Joint}}
\providecolor{raggreen}{RGB}{240,250,240}
\providecolor{flatblue}{RGB}{235,247,252}
\providecolor{structpink}{RGB}{252,238,244}
\providecolor{ourspeach}{RGB}{255,238,221}
\providecolor{trainpurple}{RGB}{244,240,252}
\providecommand{\NA}{--}
\providecommand{\AppWideTableBegin}[1][!t]{\begin{table*}[#1]}
\providecommand{\AppWideTableEnd}{\end{table*}}
\makeatletter
\@ifundefined{AppHereTable}{%
  \newenvironment{AppHereTable}{%
    \par\smallskip\noindent\begin{minipage}{\columnwidth}\centering
  }{%
    \end{minipage}\par\smallskip
  }%
}{}
\makeatother

\appendix
\setcounter{secnumdepth}{2}

\section{Complete Method Specification}
\label{app:complete_method}

\subsection{Interfaces, State, and Notation}
\label{app:interfaces}

The main text uses $F$ as functional shorthand; here we separate its frozen
roles to state the experimental assignments without ambiguity.  Let
$F_{\mathrm{ans}}$ denote the answering model, $F_{\mathrm{route}}$ the model
queried by SR-QR, and $F_{\mathrm{mem}}$ the model used for memory
construction, utility attribution, temporal-relation classification, and
metadata repair.  We use $F_{\mathrm{mem}}=F_{\mathrm{ans}}$ unless stated
otherwise, whereas $F_{\mathrm{route}}$ may be the same frozen model or a
different one.  Model names in the main result tables identify
$F_{\mathrm{ans}}$ only; Table~\ref{tab:app-models} reports the answering,
memory, embedding, and evaluation assignments, while $F_{\mathrm{route}}$ is
specified independently for each run.  The GPT-4 MC result replaces only
$F_{\mathrm{ans}}$; the frozen
SR-QR model and the learned external-memory state remain those of the source
run.  Let
$\operatorname{Eval}_{\mathrm{tr}}$ and
$\operatorname{Eval}_{\mathrm{rep}}$ denote the training-reward and
reporting evaluators.  They coincide for executable or exact-match feedback.
For LoCoMo OpenQA, they are intentionally different LLM judges.  None of the
frozen interfaces or evaluators is fine-tuned; the only neural parameters
optimized by \method{} are those of the residual router.

At interaction $t$, the core retrieval--memory state is
$s_t=(q_t,\mathcal G_t,\theta_t)$, where $q_t$ is the current query,
$\mathcal G_t=(\mathcal V_t,\mathcal E_t)$ is a persistent typed memory
graph, and $\theta_t$ parameterizes the residual router.  A node
$m\in\mathcal V_t$ stores preserved memory content $x_m$, a retrieval
description $d_m$, sparse keywords $K_m$, temporal metadata $s_m$, source
trajectory $\tau_m$, and a learned retrieval-utility value $Q_t(m)$.  Edges have type
$c\in\mathcal C=\{\Dense,\Sparse,\TimeRel\}$.  Dense and sparse edges are
reciprocal; a time edge is directed from an earlier memory to a later update
of the same event or state.

The forward interaction is
\begin{equation}
\label{eq:app-forward}
\begin{aligned}
\mathbf o_t
&=F_{\mathrm{route}}(q_t)
=\bigl(\widetilde{\mathcal Q}_t^d,
       \widetilde{\mathcal Q}_t^s,
       \mathbf p_t^0,\xi_t^0\bigr),\\
\mathbf w_t
&=\pi_{\theta}(q_t,\mathbf o_t),\\
\mathcal A_t
&=\operatorname{Retrieve}
  (\widetilde{\mathcal Q}_t^d,
   \widetilde{\mathcal Q}_t^s,
   \mathcal G_t,\mathbf w_t),
\qquad |\mathcal A_t|\leq k,\\
y_t
&\sim p_{F_{\mathrm{ans}}}
  (\cdot\mid q_t,\mathcal A_t),
\qquad
R_t=\operatorname{Eval}_{\mathrm{tr}}(y_t;\mathcal Y_t)\in[0,1].
\end{aligned}
\end{equation}
For interactive benchmarks, $y_t$ denotes the complete generated
action--observation trace together with its terminal response; for single-turn
benchmarks, it denotes the generated response.  Thus the interaction record
below has the same task-dependent scope as the ``trajectory'' in the main text.
The interaction record
$\tau_t=(q_t,\mathcal A_t,y_t,R_t)$ is consumed either by a router update or
by a memory update.  The exposed set $\mathcal A_t$ is the coupling
interface: routing determines which memories are visible, and memory updates
change the state ranked by later routing decisions.

\subsection{Structured SR-QR}
\label{app:srqr}

As the implementation instantiation of the main-text tuple, the SR-QR prompt
requires $F_{\mathrm{route}}$ to return one strict JSON
object with five fields:
\texttt{route\_prior}, \texttt{dense\_queries},
\texttt{sparse\_queries}, \texttt{keywords}, and \texttt{confidence}.
The compact tuple $\mathbf o_t$ in the main text suppresses the auxiliary keyword set
$K_t$, which is parsed from \texttt{keywords} and used only by sparse
retrieval.
After sanitization, each route contains one to three rewrites.  If exactly one
route is empty, the parser inserts the original query on that route.  Dense
rewrites preserve semantic intent and vary its expression.
Sparse rewrites are short phrases that retain identifiers, proper nouns,
error names, versions, numbers, and other exact-match cues.  The global
keyword list contains at most eight discriminative tokens.

The parser removes duplicate rewrites while preserving order, caps each list
at three, normalizes nonnegative finite prior values onto the two-dimensional
simplex, and clips confidence to $[0,1]$.  A malformed response, failed call,
invalid prior, or pair of empty rewrite lists triggers the deterministic
fallback
\begin{equation}
\label{eq:app-srqr-fallback}
\begin{aligned}
\mathbf p_t^0&=(0.5,0.5),&
\widetilde{\mathcal Q}_t^d&=\bigl\{q_t\bigr\},\\
\widetilde{\mathcal Q}_t^s&=\bigl\{q_t\bigr\},&
\xi_t^0&=0.
\end{aligned}
\end{equation}
The parser records the fallback event for diagnostics.  The fallback guarantees
a valid retrieval action and introduces no trainable parameters.

\subsection{Residual Router}
\label{app:router}

Following the parameterization in the main text, the residual router operates
on the current query and the structured SR-QR output.  Its query encoder and
policy head are
\begin{equation}
\label{eq:app-router-policy}
\begin{aligned}
\mathbf c_t
&=\phi_\psi(q_t,\mathbf o_t)\in\mathbb R^h,\\
\boldsymbol\Delta_t
&=f_\omega(\mathbf c_t)\in\mathbb R^2,\\
\bar p_t^0(a)
&=\frac{p_t^0(a)+\varepsilon}
{\sum_{a'\in\{d,s\}}(p_t^0(a')+\varepsilon)},\\
\boldsymbol\pi_t
&=\operatorname{softmax}
  \left(\log\bar{\mathbf p}_t^0+\boldsymbol\Delta_t\right).
\end{aligned}
\end{equation}
Here $\theta=(\psi,\omega)$, and $\varepsilon>0$ smooths the prior.  The final
layer of $f_\omega$ is initialized to zero, so the initial mean policy exactly
recovers the smoothed LLM prior.  During router training,
\begin{equation}
\label{eq:app-beta-policy}
\zeta_t\sim\operatorname{Beta}
(\kappa\pi_t^d,\kappa\pi_t^s),
\qquad \mathbf w_t=(\zeta_t,1-\zeta_t).
\end{equation}
During memory phases and evaluation, exploration is disabled and
$\mathbf w_t=\boldsymbol\pi_t$.

Let $B$ be an exponential moving-average baseline with retention
$\mu=0.9$, equivalently $\lambda_{\mathrm{ema}}=1-\mu=0.1$ in the main-text
notation.  At interaction $t$, the advantage is
$\widehat A_t=R_t-B_{t-1}$ and the baseline update is
$B_t=\mu B_{t-1}+(1-\mu)R_t$.  Over an optimization minibatch
$\mathcal B$, the router minimizes
\begin{equation}
\label{eq:app-router-loss}
\begin{aligned}
\widehat{\mathcal L}_{\mathrm{router}}
&=\widehat{\mathcal L}_{\mathrm{PG}}
  +\beta_{\mathrm{KL}}\widehat{\mathcal L}_{\mathrm{KL}},\\
\widehat{\mathcal L}_{\mathrm{PG}}
&=-\frac{1}{|\mathcal B|}\sum_{t\in\mathcal B}
\operatorname{sg}[\widehat A_t]\,
\log p_{\theta}(\zeta_t\mid q_t,\mathbf o_t),\\
\widehat{\mathcal L}_{\mathrm{KL}}
&=\frac{1}{|\mathcal B|}\sum_{t\in\mathcal B}
D_{\mathrm{KL}}\!\left(
\operatorname{Cat}(\boldsymbol\pi_t)
\,\middle\|\,
\operatorname{Cat}(\bar{\mathbf p}_t^0)\right).
\end{aligned}
\end{equation}
The score-function term trains through non-differentiable retrieval, while the
KL term limits unnecessary departure from the frozen prior.

\subsection{Hybrid Retrieval and Time Expansion}
\label{app:retrieval}

Every sanitized rewrite creates an independent ranked list.  Dense lists use
cosine similarity between rewrite and memory-description embeddings.  Sparse
lists use BM25 over memory keywords; when SR-QR returns an explicit keyword
set, the same set is used for all sparse rewrites.  As a bounded implementation
of the main-text candidate pool, the retriever first
requests up to
$b_k=\min\{N,\max(3k,10)\}$ candidates per rewrite, where $N$ is a
benchmark-specific candidate cap fixed by the task configuration.  Let
$\mathcal L_{t,a,i}$ be list $i$ on route $a\in\{d,s\}$, and let $N_a$
be the number of nonempty lists actually returned on route $a$.  For a
candidate pool $\mathcal P$, define the normalized route-specific
reciprocal-rank score $G_{t,a}$ and fused score $S_t$ as
\begin{equation}
\label{eq:app-hybrid-score}
\begin{aligned}
G_{t,a}(m;\mathcal P)
&=\frac{1}{N_a}
\sum_{i=1}^{N_a}
\frac{1}{\eta+\operatorname{rank}_{a,i}^t(m;\mathcal P)},\\
S_t(m;\mathcal P)
&=\sum_{a\in\{d,s\}}w_t^aG_{t,a}(m;\mathcal P)
+\frac{\lambda_Q}
{\eta+\operatorname{rank}_{Q}^t(m;\mathcal P)}.
\end{aligned}
\end{equation}
The $1/N_a$ factor prevents a route from gaining mass merely by emitting more
rewrites.  Equal $Q$ values share the same rank, avoiding arbitrary ordering
of cold-start memories.

Let $\mathcal P_t^{(0)}$ be the union of all dense and sparse rewrite lists,
and let
$\mathcal K_t=\operatorname{TopK}_k(S_t(\cdot;\mathcal P_t^{(0)}))$.
The system augments this initial pool with the incoming and outgoing
\TimeRel{} neighbors of the seeds:
\begin{equation}
\label{eq:app-temporal-expansion}
\begin{aligned}
\mathcal N_t^{\mathrm{time}}
&=\bigcup_{m\in\mathcal K_t}
\left(\mathcal N_{\mathrm{in}}^{\TimeRel}(m)
\cup\mathcal N_{\mathrm{out}}^{\TimeRel}(m)\right),\\
\mathcal P_t^{(1)}&=\mathcal P_t^{(0)}\cup
\mathcal N_t^{\mathrm{time}},\\
\mathcal A_t&=\operatorname{TopK}_k
\left(S_t(\cdot;\mathcal P_t^{(1)})\right).
\end{aligned}
\end{equation}
All route-specific ranks and $Q$ ranks are recomputed on the expanded pool;
temporal neighbors therefore compete under the same hybrid score rather than
receiving a separate additive bonus.  Only the final set $\mathcal A_t$ is
inserted into the answering context and receives exposure statistics or
direct credit.

\subsection{Relational Memory Graph}
\label{app:memory_construction}

During a memory phase, $F_{\mathrm{mem}}$ distills the complete interaction
trajectory, including unsuccessful interactions, into a reusable procedural
memory while retaining the source trajectory as provenance.  Conversation
histories are segmented into chronologically ordered user--assistant turn
pairs.  Other benchmarks produce one candidate memory per task interaction.
No prompt used for memory construction, utility attribution, or graph repair
receives a gold answer, corrected answer, or evaluator rationale.

Let $\mathcal E_c\subseteq\mathcal E_t$ denote the arcs of type
$c\in\mathcal C$.  For a new node $m$, let $\mathbf e_m$ be the embedding of
its description.  Each accepted dense or sparse pair is closed reciprocally:
write $m\overset{c}{\leftrightarrow}v$ when both $(m,v)$ and $(v,m)$ belong
to $\mathcal E_c$.  Then
\begin{equation}
\label{eq:app-edge-construction}
\begin{aligned}
m\overset{d}{\leftrightarrow}v
&\Longleftrightarrow v\in\operatorname{TopK}_d(m),\ 
\cos(\mathbf e_m,\mathbf e_v)\geq\tau_d,\\
m\overset{s}{\leftrightarrow}v
&\Longleftrightarrow \operatorname{Jac}(K_m,K_v)\geq\tau_s.
\end{aligned}
\end{equation}
Candidate temporal pairs are drawn from the dense and sparse neighborhoods.
$F_{\mathrm{mem}}$ labels each pair \TimeRel{} or
\textsc{None}; accepted edges are oriented from the earlier record to the
later continuation or revision.  Topical similarity without an update to the
same event or state is insufficient.

\subsection{Memory Valuation and Relational Credit}
\label{app:valuation}

For every exposed memory $m\in\mathcal A_t$, a deterministic utility prompt
gives $F_{\mathrm{mem}}$ that memory, the query, the task-dependent generated
trace $y_t$, and the scalar terminal reward.  It returns one attributed score
$u_t(m)\in[0,1]$.  If any required score is invalid, the complete TD update
for that interaction is skipped rather than imputed.

A newly inserted memory inherits the mean pre-update value of the context
that produced it:
\begin{equation}
\label{eq:app-new-memory-value}
\widehat Q_t(m_t^{\mathrm{new}})=
\begin{cases}
|\mathcal A_t|^{-1}\sum_{m\in\mathcal A_t}Q_t(m),
&\mathcal A_t\neq\varnothing,\\
Q_{\mathrm{init}},&\mathcal A_t=\varnothing.
\end{cases}
\end{equation}
The direct TD-style residual for an exposed memory is
\begin{equation}
\label{eq:app-memory-td}
\delta_t(m)=R_tu_t(m)
+\gamma\widehat Q_t(m_t^{\mathrm{new}})-Q_t(m).
\end{equation}
All residuals in one interaction use the same pre-update snapshot.

Credit propagates backward along incoming typed edges.  Dense and sparse
relations are reciprocal, while a directed time relation propagates credit
from a later exposed state to its earlier predecessor.  For paths of at most
$D$ edges, define the strongest coefficient
\begin{equation}
\label{eq:app-relational-credit}
c_t^{(D)}(v,m)=
\max_{\substack{p:v\rightsquigarrow m\\1\leq|p|\leq D}}
\prod_{e\in p}\rho_{c(e)},
\end{equation}
with value zero when no valid path exists.  If
$\mathcal S_t(v)=\{m\in\mathcal A_t:c_t^{(D)}(v,m)>0\}$, the effective
residual is
\begin{equation}
\label{eq:app-propagated-td}
\begin{aligned}
\bar\delta_t(v)
&=\frac{1}{|\mathcal S_t(v)|}
\sum_{m\in\mathcal S_t(v)}
c_t^{(D)}(v,m)\delta_t(m),\\
\Delta_t(v)
&=
\begin{cases}
\delta_t(v),&v\in\mathcal A_t,\\
\bar\delta_t(v),
&\substack{v\notin\mathcal A_t,\\
\mathcal S_t(v)\neq\varnothing},\\
0,&\text{otherwise}.
\end{cases}
\end{aligned}
\end{equation}
Taking a maximum within each exposed-memory traversal and a mean across
different exposed memories prevents path multiplicity and graph degree from
mechanically increasing update magnitude.  The projected update is
\begin{equation}
\label{eq:app-memory-update}
Q_{t+1}(v)=
\Pi_{[Q_{\min},Q_{\max}]}
\left(Q_t(v)+\alpha_Q\Delta_t(v)\right).
\end{equation}
Equation~\eqref{eq:app-memory-update} applies to every $v\in\mathcal V_t$.
The newly constructed node is inserted with
$Q_{t+1}(m_t^{\mathrm{new}})=\widehat Q_t(m_t^{\mathrm{new}})$ and becomes
eligible for retrieval and outcome-conditioned updates from the next
interaction onward.

\subsection{Channel-Specific Retrieval-Metadata Repair}
\label{app:repair}

Within the main text's abstract memory update $\mathcal U_{\mathrm M}$, the
implementation may additionally repair retrieval metadata from
channel-specific terminal-success statistics.  These statistics are distinct
from the LLM utility scores in Eq.~\eqref{eq:app-memory-td}.  Let
$s_t^{\mathrm{succ}}\in\{0,1\}$ denote the task evaluator's binary terminal
success decision, which is kept separate from a potentially graded reward
$R_t\in[0,1]$.  For channel $c\in\mathcal C$, let
$N_t(m,c)$ be the number of final-context exposures of $m$ through $c$, and
let $C_t(m,c)$ be the sum of $s_t^{\mathrm{succ}}$ over those exposures.
The empirical correctness ratio is
\begin{equation}
\label{eq:app-repair-trigger}
\operatorname{Acc}_t(m,c)=
\begin{cases}
C_t(m,c)/N_t(m,c),&N_t(m,c)>0,\\
1,&N_t(m,c)=0.
\end{cases}
\end{equation}
A channel is eligible when
$N_t(m,c)\geq N_{\min}$ and
$\operatorname{Acc}_t(m,c)<\vartheta_{\mathrm{heal}}$, subject to a per-memory
refresh budget.
The diagnosis prompt contains the source trajectory, current description and
keywords, channel statistics, and all incident typed neighbors.  A dense repair
regenerates the description and its dense edges; a sparse repair regenerates
keywords and sparse edges; a temporal repair revalidates incident time edges.
The preserved memory content and its learned $Q$ value are not rewritten.

\subsection{Alternating Co-Adaptation}
\label{app:schedule}

In a \RouterPhase{} phase, the graph, values, metadata, and exposure
statistics are read-only; sampled route mixtures update only $\theta$.  In a
\MemoryPhase{} phase, the router is frozen, retrieval uses its deterministic
mean, and the graph evolves sequentially.  Validation and test run inside a
fully frozen evaluation context: both states are read-only, and no prediction
or evaluator output enters later training.

For a co-adaptation run, let
$\mathsf S_{\mathrm{base}}=(z_1,\ldots,z_{N_{\mathrm{base}}})$ denote the
base schedule, where $z_1=\RouterPhase$,
$z_r\in\{\RouterPhase,\MemoryPhase\}$, and $z_{r+1}\neq z_r$.
Every phase is one complete pass over $\mathcal D_{\mathrm{tr}}$.  The
executed schedule is
\begin{equation}
\label{eq:app-schedule-construction}
\mathsf S_{\mathrm{exec}}=
\begin{cases}
(\MemoryPhase_{\mathrm{boot}})\Vert\mathsf S_{\mathrm{base}},
&\mathcal V_0=\varnothing,\\
\mathsf S_{\mathrm{base}},&\mathcal V_0\neq\varnothing.
\end{cases}
\end{equation}
We set $N_{\mathrm{ph}}=|\mathsf S_{\mathrm{exec}}|$, so the phase count in
the main-text objective includes the bootstrap pass whenever it is executed.
The subscript ``boot'' is only a bookkeeping label for an ordinary
\MemoryPhase{} update applied to an empty graph, not a third update type.  This
Appendix uses one-based schedule positions when referring to phase numbers.
Thus, an empty-graph task uses phase~1 to construct memory and begins Router
optimization in phase~2.  A history-initialized task begins directly with
Router in phase~1.  Throughout the paper, \emph{alternating} therefore means
strict switching between Router and Memory phases in co-adaptation runs, with
each phase holding its counterpart fixed.  Table~\ref{tab:app-dataset-hparams}
reports each dataset's complete executed sequence and therefore fixes both
ordering and pass count.

The simultaneous ablation uses \JointPhase{} passes, updating both components
on the same training stream.  The two-stage ablation performs one contiguous
Router block followed by one Memory block, with no return to Router
optimization.  These are experimental comparators and are not branches of the
core training algorithm.  LoCoMo-MC and LongMemEval-MC use a separate
fixed-memory protocol: their history graphs are constructed once, and all
training passes update only the router.

Algorithm~\ref{alg:app-coevo} expands the interaction-level computation
abstracted by the phase-wise training procedure in the main text.  Rather than
repeating the outer traversal over the phase schedule, it specifies the shared
retrieval pipeline and the component-specific state transition induced by one
training interaction.  The active phase is supplied by
$\mathsf S_{\mathrm{exec}}$; a
$\MemoryPhase_{\mathrm{boot}}$ phase follows the Memory branch below.

During a \RouterPhase{} phase, $\mathcal G_t$ is strictly read-only.  Each
interaction contributes a score-function loss with KL regularization to the
router buffer $\mathcal B_{\mathrm R}$.  AdamW is applied at the declared
optimizer-minibatch boundary, with any residual entries flushed at phase end.
Batching therefore changes only the timing of gradient application, not the
order in which trajectories are collected.

During a \MemoryPhase{} or bootstrap phase, the router and EMA state are held
fixed and retrieval uses the deterministic mean policy $\boldsymbol\pi_t$.
All TD residuals and propagation coefficients are computed from one pre-update
graph snapshot, after which existing-node utilities are updated synchronously.
The new node is then inserted with its inherited utility and becomes eligible
for retrieval and outcome-conditioned updates from the next interaction.
Channel-specific terminal-success statistics may subsequently trigger
budget-constrained metadata repair; repair preserves the stored memory content
and its updated utility.  Invalid model outputs follow the conservative
semantics in Section~\ref{app:failure_semantics}, preventing partial state
mutations.

\subsection{Failure Semantics and State Integrity}
\label{app:failure_semantics}

The following implementation contract makes all LLM-assisted state changes
fail conservatively.
\begin{itemize}
  \item Invalid SR-QR output falls back to the original query on both routes
  with a uniform prior.
  \item Invalid temporal classification is treated as \textsc{None}; no time
  edge is created.
  \item Invalid or incomplete utility output skips the entire TD update for
  that interaction.
  \item Invalid metadata diagnosis or repair output leaves the stored
  content, graph structure, and learned utility unchanged.
  \item Evaluation freezes both components and never feeds held-out outcomes
  back into the training state.
\end{itemize}

\begin{algorithm*}[t]
\small
\caption{Interaction-Level Retrieval and Component-Specific Updates}
\label{alg:app-coevo}
\begin{algorithmic}[1]

\Statex \textbf{Input:} Interaction $(q_t,\mathcal Y_t)$; active phase
$z_r\in\{\textsc{Router},\textsc{Memory}\}$;
router $\theta$; graph $\mathcal G_t=(\mathcal V_t,\mathcal E_t)$;
EMA state $B_{t-1}$; router-loss buffer $\mathcal B_{\mathrm R}$;
optimizer state $\Omega_t$;
frozen $F_{\mathrm{ans}},F_{\mathrm{route}},F_{\mathrm{mem}}$
\Statex \textbf{Output:} Exposure $\mathcal A_t$, answer $y_t$, reward $R_t$,
and updated $(\theta^{+},B_t,\mathcal G_{t+1},\mathcal B_{\mathrm R},
\Omega_t^{+})$

\Statex \textit{Route formation and retrieval}
\State $(\mathbf o_t,K_t)\gets\operatorname{Sanitize}
       (F_{\mathrm{route}}(q_t))$
\State $\mathbf c_t\gets\phi_\psi(q_t,\mathbf o_t)$;
\State $\boldsymbol\pi_t\gets
\operatorname{softmax}(\log\bar{\mathbf p}_t^0
+f_\omega(\mathbf c_t))$
\If{$z_r=\textsc{Router}$}
    \State $\zeta_t\sim
    \operatorname{Beta}(\kappa\pi_t^d,\kappa\pi_t^s)$;
    $\mathbf w_t\gets(\zeta_t,1-\zeta_t)$
\Else
    \State $\mathbf w_t\gets\boldsymbol\pi_t$
\EndIf
\State Form $\mathcal L_{t,d,i}$ by description-embedding similarity and
$\mathcal L_{t,s,i}$ by BM25; truncate every list at $b_k$
\State $\mathcal P_t^{(0)}\gets\bigcup_{a\in\{d,s\},i}
       \mathcal L_{t,a,i}$;
$\mathcal K_t\gets\operatorname{TopK}_k
       (S_t(\cdot;\mathcal P_t^{(0)}))$
\State $\mathcal P_t^{(1)}\gets\mathcal P_t^{(0)}\cup
       \displaystyle\bigcup_{m\in\mathcal K_t}
       \bigl(\mathcal N_{\mathrm{in}}^{\textsc{Time}}(m)
       \cup\mathcal N_{\mathrm{out}}^{\textsc{Time}}(m)\bigr)$
\State Recompute all route and $Q$ ranks on $\mathcal P_t^{(1)}$;
$\mathcal A_t\gets\operatorname{TopK}_k
       (S_t(\cdot;\mathcal P_t^{(1)}))$
\State $y_t\sim p_{F_{\mathrm{ans}}}
       (\cdot\mid q_t,\mathcal A_t)$;
$R_t\gets\operatorname{Eval}_{\mathrm{tr}}(y_t;\mathcal Y_t)$;
$s_t^{\mathrm{succ}}\gets\operatorname{Success}(y_t;\mathcal Y_t)$
\State $\tau_t\gets(q_t,\mathcal A_t,y_t,R_t)$

\Statex \textit{Component-specific update}
\If{$z_r=\textsc{Router}$}
    \State $\widehat A_t\gets R_t-B_{t-1}$;
    $B_t\gets\mu B_{t-1}+(1-\mu)R_t$
    \State $\ell_t\gets
    -\operatorname{sg}[\widehat A_t]
    \log p_\theta(\zeta_t\mid q_t,\mathbf o_t)
    +\beta_{\mathrm{KL}}D_{\mathrm{KL}}
    (\operatorname{Cat}(\boldsymbol\pi_t)
    \|\operatorname{Cat}(\bar{\mathbf p}_t^0))$
    \State Append $\ell_t$ to $\mathcal B_{\mathrm R}$
    \If{at an optimizer-minibatch or pass-end boundary}
        \State $(\theta^{+},\Omega_t^{+})\gets
        \operatorname{AdamW}(\theta,\Omega_t;
        \operatorname{mean}(\mathcal B_{\mathrm R}))$;
        $\mathcal B_{\mathrm R}\gets\varnothing$
    \Else
        \State $\theta^{+}\gets\theta$; $\Omega_t^{+}\gets\Omega_t$
    \EndIf
    \State $\mathcal G_{t+1}\gets\mathcal G_t$
\Else
    \State $\theta^{+}\gets\theta$; $\Omega_t^{+}\gets\Omega_t$;
    $B_t\gets B_{t-1}$
    \State $m_t^{\mathrm{new}}\gets
    \operatorname{Construct}_{F_{\mathrm{mem}}}(\tau_t)$
    \State $\widehat Q_t(m_t^{\mathrm{new}})\gets
    \operatorname{mean}_{m\in\mathcal A_t}Q_t(m)$,
    or $Q_{\mathrm{init}}$ if $\mathcal A_t=\varnothing$
    \ForAll{$m\in\mathcal A_t$}
        \State $u_t(m)\gets
        \operatorname{Attribute}_{F_{\mathrm{mem}}}(m,\tau_t)$
    \EndFor
    \If{all attribution outputs are valid}
        \State $\delta_t(m)\gets R_tu_t(m)
        +\gamma\widehat Q_t(m_t^{\mathrm{new}})-Q_t(m)$
        for every $m\in\mathcal A_t$
        \State Compute $c_t^{(D)}(v,m)$ and $\Delta_t(v)$ on the
        pre-update graph for every $v\in\mathcal V_t$
        \State Synchronously set $Q_{t+1}(v)\gets
        \Pi_{[Q_{\min},Q_{\max}]}
        (Q_t(v)+\alpha_Q\Delta_t(v))$ for every $v\in\mathcal V_t$
    \Else
        \State $Q_{t+1}(v)\gets Q_t(v)$ for every $v\in\mathcal V_t$
    \EndIf
    \State Insert $m_t^{\mathrm{new}}$ with value
    $\widehat Q_t(m_t^{\mathrm{new}})$ and provenance $\tau_t$
    \State Add reciprocal \textsc{Dense}/\textsc{Sparse} edges and
    verified, directed \textsc{Time} edges
    \State Update channel-specific statistics of $\mathcal A_t$ using
    $s_t^{\mathrm{succ}}$
    \ForAll{eligible $(m,c)$ satisfying
    $N_t(m,c)\geq N_{\min}$ and
    $\operatorname{Acc}_t(m,c)<\vartheta_{\mathrm{heal}}$}
        \State Repair metadata channel $c$ using $F_{\mathrm{mem}}$;
        preserve $x_m$ and $Q_{t+1}(m)$
    \EndFor
    \State Let $\mathcal G_{t+1}$ be the resulting graph
\EndIf
\State \Return $(\mathcal A_t,y_t,R_t,\theta^{+},B_t,
\mathcal G_{t+1},\mathcal B_{\mathrm R},\Omega_t^{+})$

\end{algorithmic}
\end{algorithm*}
\clearpage

\section{Experimental Protocols}
\label{app:experimental_protocols}

\subsection{Benchmarks and Metrics}
\label{app:benchmarks}

We evaluate seven benchmark families with their task-native interfaces.  Each
reported unit is held out, and evaluation outcomes never update the router or
memory state.
\begin{itemize}
  \setlength{\itemsep}{1pt}
  \setlength{\parsep}{0pt}
  \setlength{\topsep}{2pt}
  \item \textbf{LifelongAgentBench (LLAB), OS interaction} evaluates
  multi-step operating-system planning across sessions.  Tasks run in the
  benchmark environment, with terminal-state success and the MemQ cap of 15
  agent steps.
  \item \textbf{LiveCodeBench v6} evaluates competitive-programming
  generation: one program is executed against the benchmark tests per problem,
  and we report Pass@1.
  \item \textbf{MMMU Pro} evaluates multiple-choice multimodal reasoning with
  exact-choice accuracy.
  \item \textbf{GPQA Diamond} evaluates graduate-level physics, chemistry, and
  biology questions with exact multiple-choice accuracy.
  \item \textbf{BFCL} evaluates multi-turn function calling, including API
  selection, argument construction, and recovery from tool-side errors.
  Evaluation uses the benchmark parser and backend-state comparison.
  \item \textbf{LoCoMo} is evaluated as both free-form OpenQA and MC-10.
  OpenQA excludes category~5 adversarial questions before splitting and reports
  overall, multi-hop, temporal, open-domain, and single-hop judged accuracy;
  MC-10 uses exact-choice accuracy.
  \item \textbf{LongMemEval-MC} converts each complete history--question bundle
  into a 10-choice task scored by exact-choice accuracy.
\end{itemize}

Executable and exact-choice tasks use task-native correctness for both reward
and reporting.  Only LoCoMo OpenQA separates these roles: GPT-4o-mini supplies
training rewards, while DeepSeek-V4-Pro judges frozen test predictions and
never enters training or memory updates.

\subsection{Dataset Construction and Partitions}
\label{app:data_splits}

Conversational protocols hold out different units: LoCoMo-MC splits complete
conversations, LoCoMo OpenQA splits QA pairs within each conversation, and
LongMemEval-MC assigns each complete history--question bundle to one
conversation-disjoint partition.  The first five task-oriented benchmarks use
the counts and \emph{valid}/\emph{test} terminology of MemQ Appendix~D.

\AppWideTableBegin
\centering
\begin{threeparttable}
\scriptsize
\setlength{\tabcolsep}{4pt}
\renewcommand{\arraystretch}{1.08}
\begin{tabularx}{\textwidth}{@{}l X
>{\centering\arraybackslash}p{0.13\textwidth}
>{\centering\arraybackslash}p{0.13\textwidth}
>{\centering\arraybackslash}p{0.13\textwidth}@{}}
\toprule
\textbf{Benchmark} & \textbf{Partition unit} & \textbf{Train} &
\textbf{Validation} & \textbf{Test} \\
\midrule
LLAB OS Interaction & OS task & 500 (full) & \NA & Official held-out \\
LiveCodeBench v6 & Programming problem & 140 & 35 & \NA \\
MMMU Pro & Multimodal question & 1,384 & 346 & \NA \\
BFCL & Function-call task & 400 & \NA & 100 \\
GPQA Diamond & Multiple-choice question & 158 & \NA & 40 \\
\midrule
LoCoMo OpenQA & QA pair within each conversation & 60\% & 20\% & 20\% \\
LoCoMo-MC & Entire conversation & 60\% (6) & 20\% (2) & 20\% (2) \\
LongMemEval-MC & Entire conversation / history--question bundle & 60\% (300) &
20\% (100) & 20\% (100) \\
\bottomrule
\end{tabularx}
\caption{Dataset partitions and split units.}
\label{tab:app-splits}
\begin{tablenotes}
\footnotesize
\item A dash indicates that the adopted protocol does not define a separate
split with that name.  LoCoMo-MC uses a conversation-disjoint 6:2:2 split.
LoCoMo OpenQA removes category~5 and then performs a question-type-stratified
60/20/20 split inside every conversation; the main OpenQA result averages five
independent within-conversation splits.  LongMemEval-MC uses a
conversation-disjoint 300/100/100 split (6:2:2), without separating a bundle's
history from its question.
\end{tablenotes}
\end{threeparttable}
\AppWideTableEnd

Graph initialization is benchmark-specific.  LoCoMo-MC and LongMemEval-MC
convert each supplied history to a fixed graph before QA, so rewards update
only the router.  LoCoMo OpenQA initializes a graph from each complete
conversation, after which only training questions may update it or the router.
The five task-oriented benchmarks instead construct an initially empty memory
from training trajectories during Memory phases.  Validation selects a
checkpoint only when a disjoint test split also exists.  A sole held-out split,
even if named \emph{valid}, is reserved for one-time reporting from the final
scheduled state; the same rule applies when no validation split exists.  The
selected state is frozen for held-out evaluation, whose predictions, rewards,
and evaluator outputs are never fed back into training.

\subsection{Baselines and Reproduction Controls}
\label{app:baselines}

We group baselines by the state exposed to the answering model.
\textbf{Context and retrieval baselines} are No Memory, Full Context, RAG
(embedding retrieval without learned utility reranking), Self-RAG (relevance,
support, and utility critiques), and graph-enhanced LightRAG.  Under the MemQ
protocol, Self-RAG uses the common frozen answering backbone rather than a
task-specific fine-tuned critique model.

\textbf{External and structured memory baselines} are Mem0, LangMem, MemoryOS,
MemP, A-MEM, Zep, Memobase, and StructMem, each retaining its native memory
construction and organization.  \textbf{Adaptive and model-integrated
baselines} are MemRL and MemQ, which learn external-memory utilities with a
frozen answering backbone, plus MemGen, AutoCompressor, and ElasticMem, whose
state is more directly coupled to that backbone.

For LLAB, LiveCodeBench, MMMU Pro, GPQA Diamond, and BFCL, we follow the MemQ
comparison protocol: RAG and MemP use the MemRL evaluation-stack
implementations, Mem0 and MemRL use their official implementations, and
Self-RAG uses the frozen-backbone reproduction above.  A matched rerun fixes
the benchmark input, answering backbone, evaluator, and decoding while
preserving the baseline's native memory mechanism.  Retrieval and context
limits are benchmark-specific but fixed within each matched comparison.
Imported values are system-level literature references, not paired reruns.

Accordingly, the \textsc{OpenAI} and \textsc{FullContext} LoCoMo OpenQA rows
retain their reported evaluation settings and are not matched \method{} reruns;
other published OpenQA systems are interpreted likewise unless marked matched.
For MC-10, Qwen2.5-3B-Instruct and Qwen2.5-7B-Instruct define the
matched-backbone settings, whereas GPT-4 is an answering-backbone transfer
reference.

The main-text ablations are operationally fixed.  \textbf{w/o SR-QR} sends the
original question once to each route and fixes fusion to $(0.5,0.5)$.
\textbf{w/o Memory Evolution} uses independent items and direct
outcome-conditioned utility updates, disabling typed relations and relational
propagation.  Simultaneous and two-stage variants change only the Router/Memory
activation order.  All four preserve the full run's example order and
component-update budget.

\subsection{Model, Retriever, and Evaluator Assignments}
\label{app:model_matrix}

Main-table model labels denote $F_{\mathrm{ans}}$ only.  The separately
specified, frozen SR-QR model $F_{\mathrm{route}}$ may differ from
$F_{\mathrm{ans}}$ but is fixed within a benchmark run; this protocol choice is
not an additional model-comparison experiment.  Unless noted,
$F_{\mathrm{mem}}=F_{\mathrm{ans}}$.  Rewriting and answering may therefore use
different families or scales.  They communicate only through structured SR-QR
outputs and the trainable residual router, and neither frozen model is
fine-tuned.

\AppWideTableBegin
\centering
\scriptsize
\setlength{\tabcolsep}{3.5pt}
\renewcommand{\arraystretch}{1.06}
\begin{tabularx}{\textwidth}{@{}l X X X X@{}}
\toprule
\textbf{Benchmark} & \textbf{$F_{\mathrm{ans}}=F_{\mathrm{mem}}$} &
\textbf{Embedding model} & \textbf{Training reward} &
\textbf{Reporting metric/evaluator} \\
\midrule
LLAB & GPT-4o-mini & text-embedding-3-large ($d=3072$) & Environment success & Same \\
LiveCodeBench & Gemma-4-E4B-it & Qwen3-Embedding-8B & Executable tests & Pass@1 \\
MMMU Pro & Gemma-4-E4B-it & Qwen3-Embedding-8B & Exact choice & Exact choice \\
GPQA Diamond & Gemma-4-E4B-it & Qwen3-Embedding-8B & Exact choice & Exact choice \\
BFCL & Qwen3.5-35B-A3B & Qwen3-Embedding-8B & BFCL execution & Same \\
LoCoMo OpenQA & GPT-4o-mini & text-embedding-3-small & GPT-4o-mini judge & DeepSeek-V4-Pro \\
LoCoMo-MC & Qwen2.5-3B/7B-Instruct & Qwen3-Embedding-8B & Exact choice & Exact choice \\
LongMemEval-MC & Qwen2.5-3B/7B-Instruct & Qwen3-Embedding-8B & Exact choice & Exact choice \\
\bottomrule
\end{tabularx}
\caption{Frozen answering, memory, embedding, and evaluation roles used for
the main results.  The independently specified SR-QR model is not inferred
from the answering-backbone label.}
\label{tab:app-models}
\AppWideTableEnd

\paragraph{Answering-backbone transfer on MC.}
The MC graph, retrieval metadata, and learned query-side router are external to
the answering backbone.  The GPT-4 setting replaces only $F_{\mathrm{ans}}$;
the fixed graph, router, $F_{\mathrm{route}}$, embedding, retrieval, and
decoding settings remain those of the source MC run.  Because
$F_{\mathrm{mem}}$ is not invoked in fixed-graph evaluation, this row is an
answering-backbone transfer reference rather than a matched-compute baseline.

\subsection{Optimization, Initialization, and Phase Schedules}
\label{app:hyperparameters}

Only the residual router is optimized.  AdamW uses zero weight decay and
gradient-norm clipping at $1.0$; interaction losses are applied at each
benchmark-specific minibatch boundary, with a partial buffer flushed after a
Router phase.  EMA retention is $\mu=0.9$, i.e.,
$\lambda_{\mathrm{ema}}=0.1$ under the main-text update convention.  A
zero-initialized output layer makes the initial mean policy equal the smoothed
frozen SR-QR prior.  Defaults are learning rate $3\times10^{-4}$, minibatch 10,
and 5 warmup steps; MMMU Pro uses $10^{-3}$, 20, and 10, respectively.

\begin{table}[t]
\centering
\scriptsize
\setlength{\tabcolsep}{3pt}
\renewcommand{\arraystretch}{1.08}
\begin{tabularx}{\columnwidth}{@{}X c X c@{}}
\toprule
\textbf{Routing / retrieval} & \textbf{Value} &
\textbf{Memory / graph} & \textbf{Value} \\
\midrule
Residual-logit dimension & $2$ & Memory rate $\alpha_Q$ & $0.3$ \\
Rewrite cap (dense/sparse) & $3/3$ & Bootstrap discount $\gamma$ & $0.35$ \\
Beta concentration $\kappa$ & $20$ & Value interval & $[-1,5]$ \\
KL coefficient $\beta_{\mathrm{KL}}$ & $0.1$ & Propagation depth $D$ & $4$ \\
Gradient-norm clip & $1.0$ & Dense threshold $\tau_d$ & $0.6$ \\
AdamW weight decay & $0$ & Sparse threshold $\tau_s$ & $0.3$ \\
EMA retention $\mu$ & $0.9$ & Dense graph candidates & $5$ \\
RRF constant $\eta$ & $60$ & Edge decay $(\rho_d,\rho_s,\rho_t)$ & $(.8,.5,.6)$ \\
Utility-rank weight $\lambda_Q$ & $0.15$ & New-node utility prior & Context mean / neutral \\
\bottomrule
\end{tabularx}
\caption{Shared optimization, retrieval, and graph-update settings.}
\label{tab:app-shared-hparams}
\end{table}

Candidate and final exposure depths are benchmark-specific, reflecting
modality, history, and context constraints; no single retained-memory count is
imposed across datasets.  Within a benchmark the retrieval depth is fixed
across matched comparisons, subject only to a smaller candidate pool.

\begin{table}[H]
\centering
\scriptsize
\setlength{\tabcolsep}{3pt}
\renewcommand{\arraystretch}{1.06}
\begin{tabularx}{\columnwidth}{@{}>{\raggedright\arraybackslash}p{0.38\columnwidth} c X@{}}
\toprule
\textbf{Benchmark(s)} & \makecell{\textbf{Post-bootstrap}\\\textbf{passes}} &
\textbf{Executed phase sequence} \\
\midrule
LLAB, LiveCodeBench, GPQA Diamond, BFCL & 3 &
$M_{\mathrm{boot}}\Vert(R,M,R)$ \\
MMMU Pro & 16 & $M_{\mathrm{boot}}\Vert(R,M)^{\times8}$ \\
LoCoMo OpenQA & 15 & $(R,M)^{\times7}\Vert R$ \\
LoCoMo-MC & 3 & $R^{\times3}$; fixed graph \\
LongMemEval-MC & 5 & $R^{\times5}$; fixed graph \\
\bottomrule
\end{tabularx}
\caption{Per-benchmark post-bootstrap passes and complete executed phase
schedules.}
\label{tab:app-dataset-hparams}
\end{table}

$R$ and $M$ are full Router and Memory passes, $\Vert$ denotes concatenation,
and a superscript gives sequence repetition.  On empty-graph benchmarks,
$M_{\mathrm{boot}}$ constructs the graph before the listed alternating
schedule, so Router begins at phase~2 and the pass count excludes bootstrap.
History-initialized co-adaptation begins with Router; the fixed-graph MC
protocols instead remain router-only.  Minibatching changes Router-update
timing but never interaction order.

Metadata repair is restricted to evolving-memory settings: LoCoMo OpenQA uses
$(N_{\min},\vartheta_{\mathrm{heal}},B_{\mathrm{refresh}})=(15,0.15,2)$ and
LLAB, LiveCodeBench, GPQA Diamond, and BFCL use $(20,0.10,1)$.  It is disabled
for MMMU Pro and both fixed-graph MC protocols.  Repair may change
channel-specific metadata or incident relations, never preserved content or
learned utility.

\subsection{Evaluation and Decoding}
\label{app:decoding}

All auxiliary LLM calls---SR-QR, memory construction, utility attribution,
temporal classification, and metadata repair---are deterministic.  Validation
and test also disable exploration, use policy mean $\boldsymbol\pi_t$, and
generate once per instance.  Benchmark parsers score choice and executable
tasks; unparseable outputs are incorrect.

Answer generation is deterministic for LoCoMo, LongMemEval, BFCL,
LiveCodeBench, LLAB, and MMMU.  GPQA alone trains with temperature $1.0$ and
top-$p=0.95$, then uses temperature zero for validation and test.  LiveCodeBench
uses a 10,000-token cap and six-second test timeout; BFCL defaults to its
non-thinking template while also evaluating the thinking template.

Both LoCoMo OpenQA judges use fixed prompts and structured outputs: the reward
judge sees only training examples, whereas the reporting judge sees frozen
test predictions and never enters learning.  Both MC protocols use exact-choice
scoring without an LLM judge.

Aggregation follows the main tables: repeated runs with dispersion report mean
$\pm$ standard deviation, LoCoMo OpenQA aggregates five fixed
within-conversation splits, and other entries report the displayed task-level
aggregate.

\subsection{Hardware and Reproducibility}
\label{app:hardware}

All experiments ran on one server with \textbf{eight NVIDIA H20 GPUs}.  Local
frozen models used OpenAI-compatible vLLM endpoints; hosted answering,
embedding, and judge models used their configured APIs.  Gradients are limited
to residual-router parameters $\theta=(\psi,\omega)$; memory construction,
relation maintenance, value propagation, and metadata repair are nonparametric.

Within a run, model interfaces and splits remain fixed.  Random state controls
training order, router initialization, and Beta sampling; validation and test
are deterministic.  Graph and router checkpoints are serialized and restored
together for held-out evaluation.

\subsection{Limitations and Future Work}
\label{app:limitations}
\paragraph{Model-role assignments.}
Our evaluation adopts a controlled model-assignment protocol: within each
benchmark, the answering model $F_{\mathrm{ans}}$ and the SR-QR model
$F_{\mathrm{route}}$ remain fixed throughout training and evaluation, and their
pairing is not selected using held-out performance.  This design isolates the
effect of retrieval--memory co-adaptation from confounding changes in backbone
capacity or rewriting quality.  Importantly, the two roles are
interface-decoupled and may be instantiated by different frozen models:
$F_{\mathrm{route}}$ produces structured retrieval queries and routing priors,
whereas $F_{\mathrm{ans}}$ generates the task response from the retrieved
context.  The present experiments therefore evaluate \method{} under fixed,
reproducible assignments rather than exhaustively optimizing this pairing.
Future work may compare a compact set of homogeneous and heterogeneous
$(F_{\mathrm{ans}},F_{\mathrm{route}})$ configurations while holding the data
splits, prompts, embedding model, retrieval budget, training schedule, and
evaluator fixed.  Such a study would characterize the resulting
accuracy--efficiency trade-offs without conflating model assignment with the
contribution of the proposed learning framework.

\paragraph{Multi-agent memory sharing.}
The separation between retrieval policy and persistent memory also provides a
natural foundation for multi-agent memory sharing.  Multiple agents could
maintain agent-specific routers while reading from and contributing to a shared
relational memory graph, allowing useful experiences acquired by one agent to
support subsequent decisions by others.  A principled extension should preserve
the provenance and ownership of each memory, distinguish private from shared
state, and resolve redundant or conflicting updates before they affect future
retrieval.  Cross-agent credit assignment is particularly important: task
feedback should update not only the retrieving agent's policy, but also the
shared memories and relations that supported its decision, without propagating
spurious credit across unrelated agent trajectories.  Evaluating such systems
would require measuring both positive knowledge transfer and negative
interference, together with communication, storage, and synchronization costs.
A complementary theoretical direction is to characterize the stability of the
alternating optimization procedure under asynchronous writes, delayed feedback,
and multiple concurrently evolving retrieval policies.


\end{document}